\documentclass[pdflatex,sn-nature]{sn-jnl}

\usepackage{graphicx}%
\usepackage{multirow}%
\usepackage{amsmath,amssymb,amsfonts}%
\usepackage{amsthm}%
\usepackage{mathrsfs}%
\usepackage[title]{appendix}%
\usepackage{xcolor}%
\usepackage{textcomp}%
\usepackage{manyfoot}%
\usepackage{booktabs}%
\usepackage{subcaption}
\usepackage{algorithm}%
\usepackage{algorithmicx}%
\usepackage{algpseudocode}%
\usepackage{listings}%
\usepackage[table]{xcolor}

\usepackage{soul}
\usepackage{caption}
\usepackage{float}

\usepackage[switch]{lineno} 
\usepackage{pdfpages}

\theoremstyle{thmstyleone}%
\theoremstyle{thmstyletwo}%

\theoremstyle{thmstylethree}%

\begin{document}


\title[Article Title]{First On-Orbit Demonstration of a Geospatial Foundation Model}


\author*[1]{\fnm{Andrew} \sur{Du}}\email{andrew.du@adelaide.edu.au}

\author[3]{\fnm{Roberto} \sur{Del Prete}}\email{Roberto.DelPrete@esa.int}

\author[4]{\fnm{Alejandro} \sur{Mousist}}\email{Alejandro.Mousist@thalesaleniaspace.com}

\author[2]{\fnm{Nick} \sur{Manser}}\email{nick.manser@smartsatcrc.com}

\author[2]{\fnm{Fabrice} \sur{Marre}}\email{fabrice.marre@smartsatcrc.com}

\author[2]{\fnm{Andrew} \sur{Barton}}\email{andrew.barton@smartsatcrc.com}

\author[2]{\fnm{Carl} \sur{Seubert}}\email{carl.seubert@smartsatcrc.com}

\author[3]{\fnm{Gabriele} \sur{Meoni}}\email{Gabriele.Meoni@esa.int}

\author[1]{\fnm{Tat-Jun} \sur{Chin}}\email{tat-jun.chin@adelaide.edu.au}

\affil*[1]{\orgdiv{AI for Space Group}, \orgname{The University of Adelaide}, \orgaddress{\city{Adelaide}, \postcode{5000}, \state{South Australia}, \country{Australia}}}

\affil[2]{\orgname{SmartSat CRC}, \orgaddress{\city{Adelaide}, \postcode{5000}, \state{South Australia}, \country{Australia}}}

\affil[3]{\orgdiv{$\Phi$-lab}, \orgname{European Space Agency}, \orgaddress{\city{Frascati}, \postcode{00044}, \country{Italy}}}

\affil[4]{\orgname{Thales Alenia Space}, \orgaddress{\city{Cannes}, \postcode{06150}, \country{France}}}


\abstract{Geospatial foundation models (GeoFMs) promise broad generalisation capacity for Earth observation (EO) tasks, particularly under data-limited conditions. However, their large size poses a barrier to deployment on resource-constrained space hardware. To address this, we present compact variants of a Vision Transformer (ViT)-based GeoFM that preserve downstream task performance while enabling onboard execution. Evaluation across five downstream tasks and validation in two representative flight environments show that model compression and domain adaptation are critical to reducing size and resource demands while maintaining high performance under operational conditions. We further demonstrate reliable on-orbit inference with the IMAGIN-e payload aboard the International Space Station. These results establish a pathway from large GeoFMs to flight-ready, resource-efficient deployments, expanding the feasibility of onboard AI for EO missions.} 




\maketitle


\section{Introduction}\label{intro} 
Understanding Earth's dynamic systems through satellite imagery is critical for addressing a wide range of environmental and societal challenges, such as monitoring climate change~\cite{lenton2024remotely, cael2023global}, managing natural resources~\cite{hansen2013high, zeng2023monitoring}, supporting sustainable infrastructure~\cite{tay2022sea, liu2020high, ao2024national}, and enabling timely responses to natural disasters~\cite{tellman2021satellite, ainscoe2025earthquake}. To meet these needs, an increasing number of Earth observation (EO) satellites have been launched or are planned for launch, equipped with ever-advancing imagers capable of capturing high-resolution multispectral and hyperspectral imagery across a broad swath of the electromagnetic spectrum. However, the primary challenge in EO has shifted from acquiring data to efficiently analysing and extracting actionable insights from the vast volumes collected, particularly in bandwidth-limited or real-time scenarios. As a result, there is growing interest in deploying machine learning (ML) techniques, particularly deep neural networks (DNNs), directly onboard satellites to enable more advanced processing and analysis in orbit. Recent advances in space-qualified hardware accelerators, such as Intel's Myriad Vision Processing Unit (VPU) series~\cite{esa-phisat1,sasic_sasat1_space_services,mateo2023orbit, ruzicka2023fast}, Ubotica's XE platforms~\cite{munoz2024mantis,melega2023implementation,rijlaarsdam2024next}, and NVIDIA's Jetson modules~\cite{imig2023edgeai,schottl2024real, schwarz2025early}, have begun to make this feasible, opening the door to more capable and intelligent EO systems. 

To date, several EO missions have demonstrated or plan to demonstrate the feasibility of deploying ML onboard satellites. $\Phi$-sat-1~\cite{esa-phisat1,giuffrida2022phisat1} marked the first use of a DNN on a space-qualified artificial intelligence (AI) chip (i.e., Intel Movidius Myriad 2), executing a convolutional neural network (CNN) in orbit to discard cloudy images. Building on this, $\Phi$sat-2~\cite{esa-phisat2,marin2021phi} plans to support multiple onboard applications, including cloud detection, marine vessel tracking, image compression, anomaly detection, and wildfire detection, though these have yet to be demonstrated. ION-SCV 003~\cite{mateo2023orbit} showed that models can be iteratively updated by capturing images onboard, downlinking them for labelling, retraining on the ground, and uplinking the revised weights. Its successor, ION-SCV 004~\cite{ruzicka2023fast}, and SONATE-2~\cite{schwarz2025early} extended this capability by supporting onboard training on preloaded datasets acquired from previous missions. More recently, CogniSAT-6~\cite{rijlaarsdam2024next,rijlaarsdam2023autonomous} became the first EO satellite to integrate onboard AI processing, autonomous task scheduling, and near real-time insight delivery via inter-satellite links (ISL). It has also executed over 20 onboard applications including flood detection, land cover classification, and volcano monitoring~\cite{chien2024leveraging, zilberstein2024demonstrating}. Separately, ISS Mounted Accessible Global Imaging Nod-e (IMAGIN-e)~\cite{thales2024imagine,callejo2023imagin} has enabled AI processing capabilities aboard the International Space Station (ISS), serving as a testbed for edge computing in space without the need for a dedicated satellite. 


While these missions demonstrate meaningful progress, they largely rely on lightweight, task-specific models---typically CNNs---due to three practical constraints: limited onboard compute, memory, and power; scarce labelled data; and restricted uplink bandwidth, which makes updating large models impractical. Although effective, such models represent an early stage of AI adoption in space. A more scalable and transformative direction is offered by geospatial foundation models (GeoFMs)~\cite{lu2025vision}, large-scale pretrained models that support a wide range of tasks. GeoFMs provide several advantages: a single backbone can power multiple applications; far fewer labelled samples are needed to fine-tune new and existing tasks; and only small task-specific heads must be uplinked, substantially reducing bandwidth requirements.

On the ground, these foundation models have already set new benchmarks across a diverse range of EO tasks, while also demonstrating strong generalisation and data efficiency. However, most existing GeoFMs are designed with abundant resources in mind, making them ill-suited for direct deployment on satellites. Moreover, they are typically pretrained on data captured from previous EO missions, which may introduce a problem called domain gap~\cite{kouw2019introduction}---where training data (i.e., source-domain data) and operational data from a new mission or sensor (i.e., target-domain data), though seemingly similar, differ significantly in their underlying distributions. This gap often results in degraded model performance in operational settings. Achieving both efficiency and generalisation is therefore critical to realise the full potential of GeoFMs in enabling more autonomous and versatile EO missions. Efficiency is also highlighted as a critical challenge in a recent \textit{Nature Machine Intelligence} editorial~\cite{natmi2025_responsible_geofm}, particularly as geo-specific foundation models grow in size and complexity and are deployed in real-time monitoring scenarios with limited computational capacity. This challenge is even more pressing for spaceborne platforms, where strict compute and power budgets make resource-efficient adaptation essential for the practical adoption of foundation models on orbit. 

\begin{figure}[H]
\centering

\begin{subfigure}{0.95\linewidth}
    \centering
    \includegraphics[width=\linewidth]{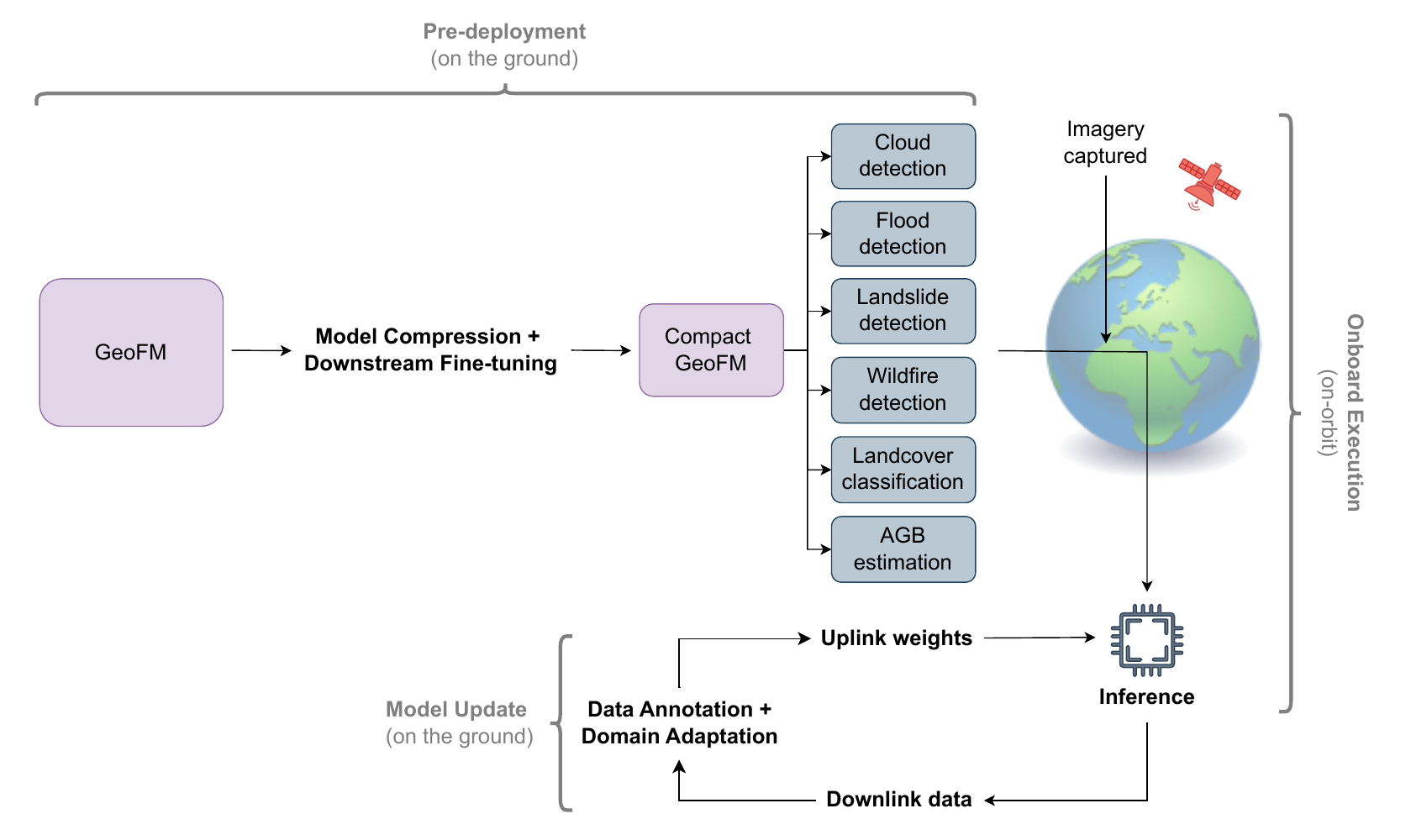}
    \subcaption{}
    \label{fig:gfm-framework}
\end{subfigure}

\vspace{1.0em}


\begin{subfigure}{0.35\linewidth}
    \centering
    \includegraphics[width=\linewidth, height=3.0cm]{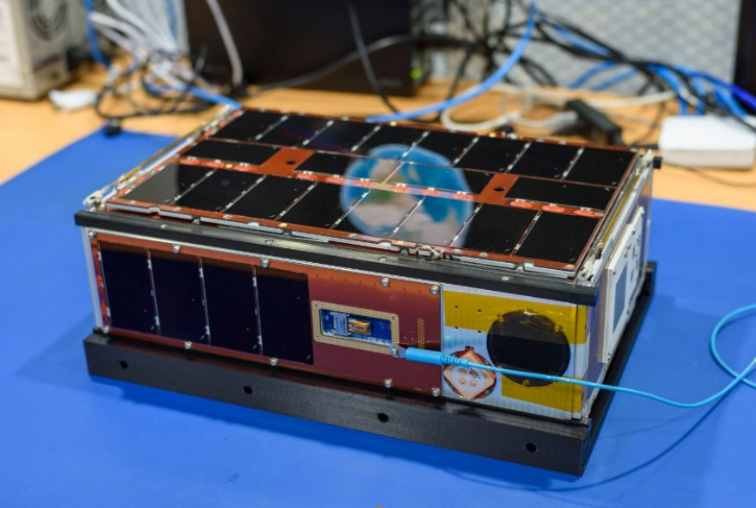}
    \subcaption{}
    \label{fig:kanyini-a}
\end{subfigure}
\hspace{0.08\linewidth}
\begin{subfigure}{0.35\linewidth}
    \centering
    \includegraphics[width=\linewidth, height=3.0cm]{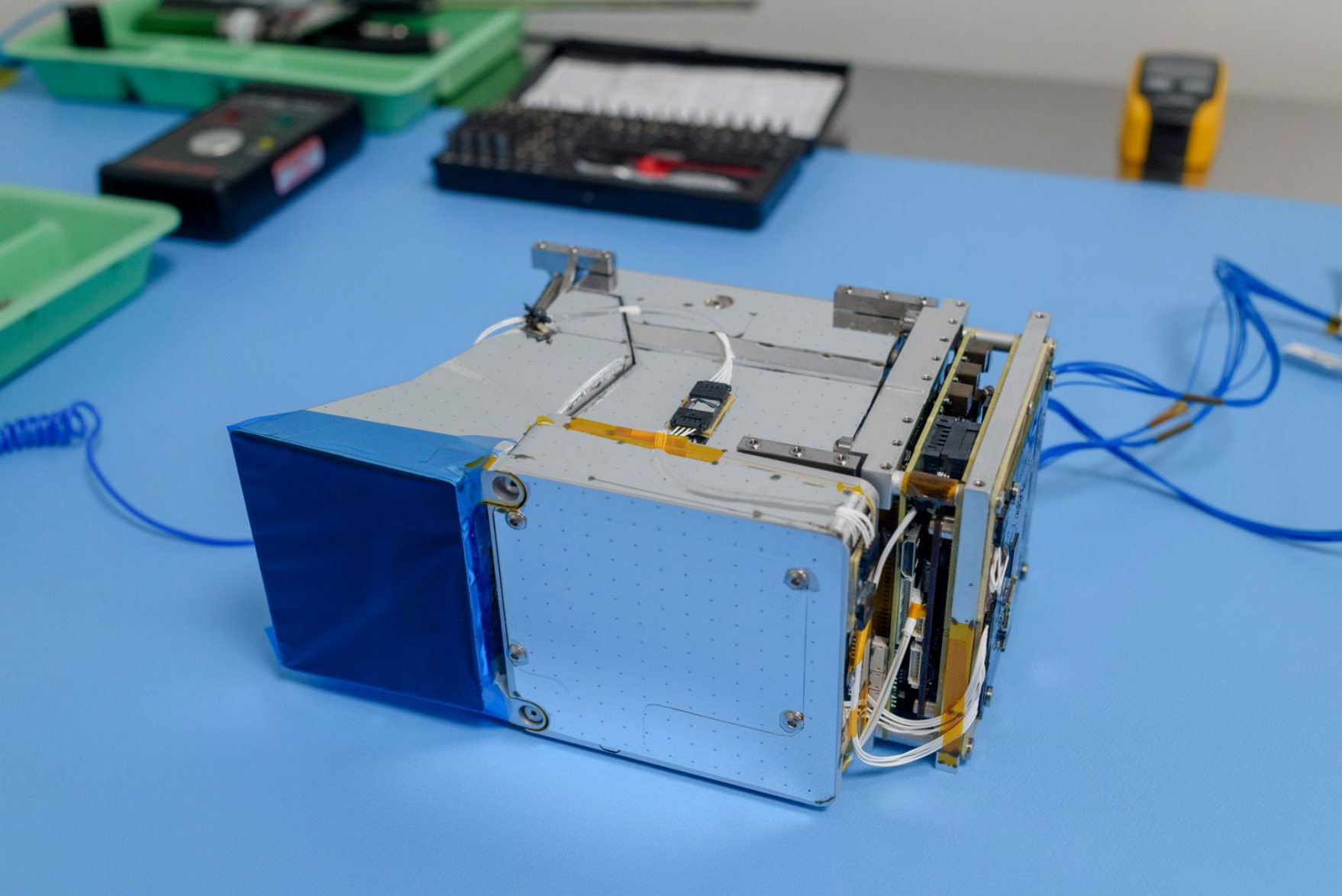}
    \subcaption{}
    \label{fig:kanyini-b}
\end{subfigure}

\begin{subfigure}{0.35\linewidth}
    \centering
    \includegraphics[width=\linewidth, height=3.0cm]{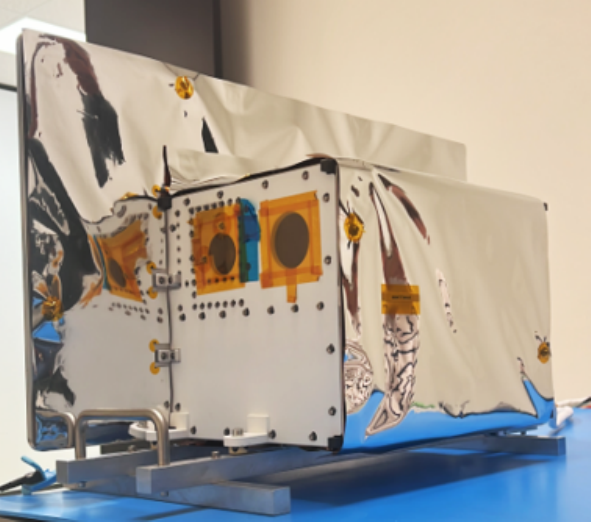}
    \subcaption{}
    \label{fig:imagin-e-payload}
\end{subfigure}
\hspace{0.08\linewidth}
\begin{subfigure}{0.35\linewidth}
    \centering
    \includegraphics[width=\linewidth, height=3.0cm]{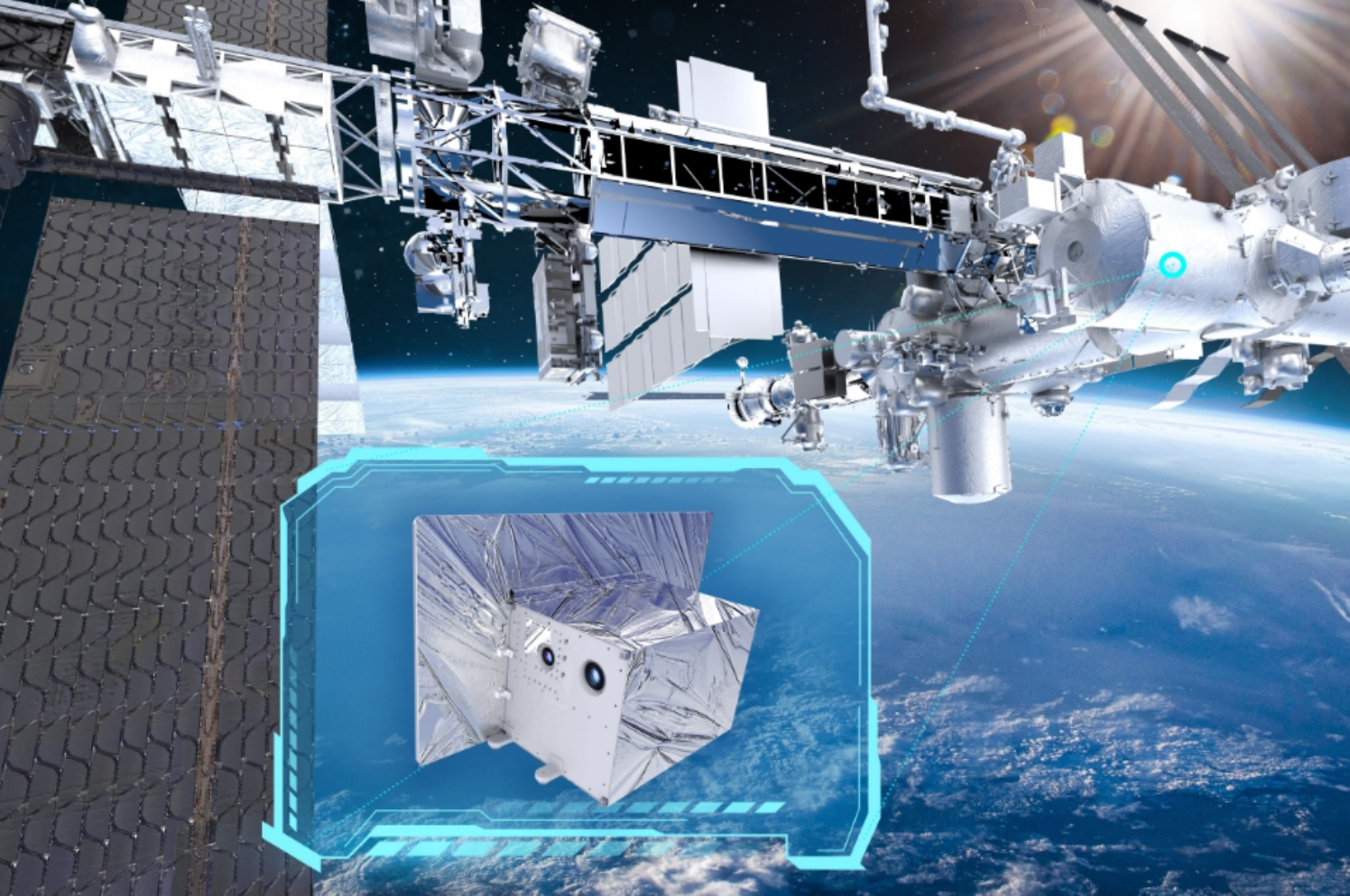}
    \subcaption{}
    \label{fig:imagin-e-iss}
\end{subfigure}

\caption{
Overview of the onboard GeoFM deployment pipeline and the satellite platforms used in this study.
\textbf{(a)} A pretrained GeoFM is compressed, paired with task heads, and deployed for onboard inference. Downlinked imagery supports continual domain adaptation, with updated weights uplinked during operations.
\textbf{(b)} Kanyini flight model and \textbf{(c)} its EO payload, HyperScout-2.
\textbf{(d)} IMAGIN-e payload and \textbf{(e)} its mounting location on the ISS.
}
\label{fig:overview}
\end{figure}

In this work, we introduce a practical framework (Fig.~\ref{fig:gfm-framework}) for enabling GeoFMs on resource-constrained EO missions. The framework balances two demands: the efficiency required by space-qualified hardware and the generalisation needed for reliable performance in orbit. By combining model compression with task-specific domain adaptation, we make GeoFMs feasible for onboard execution while preserving downstream task performance. Our contributions are:
\begin{enumerate}
    \item The first onboard deployment and execution of a GeoFM, demonstrating that such models can operate within strict satellite resource limits.
    \item A compression strategy that reduces a large GeoFM to a compact variant suitable for onboard deployment without sacrificing performance.
    \item A demonstration of effective domain adaptation under real mission conditions, bridging the gap between training data from past missions and the distribution shifts encountered in orbit.
\end{enumerate}
To validate this framework, we conducted hardware-in-the-loop evaluations on two representative platforms, with further mission details provided in Section~\ref{method0}. On Kanyini (Fig.~\ref{fig:kanyini-a}), direct on-orbit testing was not possible due to a technical anomaly affecting the EO payload interface (Fig.~\ref{fig:kanyini-b}). Instead, we deployed the GeoFM on the HyperScout-2~\cite{esposito2019hyperscout} engineering model, which replicates the flight imaging sensor and onboard processor, to demonstrate feasibility on its actual hardware architecture. On IMAGIN-e (Fig.~\ref{fig:imagin-e-payload}--\ref{fig:imagin-e-iss}), the imaging and compute subsystems are distinct, and a fault in the imager prevented live acquisitions. To address this, we uploaded a curated test set of representative scenes to the onboard compute module and executed the GeoFM directly to evaluate inference under genuine orbital conditions. Together, these evaluations highlight the practical challenges of real-world onboard deployment and show that our framework remains effective despite mission-specific constraints. Although our experiments focus on these missions, the proposed framework is general and applicable to future satellites facing similar power, memory, and compute constraints. 

\section{Results}\label{results} 

\subsection{Compact foundation models retain task performance while enabling onboard deployment}\label{results-compression} 
A key question in deploying GeoFMs onto EO satellites is whether large pretrained models can be compressed for onboard use without compromising performance. To investigate this, we applied knowledge distillation~\cite{hinton2015distilling} to Prithvi-EO-2.0-300M~\cite{szwarcman2024prithvi}, a large ViT-based geospatial foundation model, using it as the teacher for compact student variants with reduced patch embedding dimensions. In the 256-MAE-D variant, for example, the embedding dimension is reduced from 1024 to 256, yielding a $16\times$ smaller footprint (1.2~GB~$\rightarrow$~73~MB). The student is pretrained using a masked autoencoder (MAE) objective consistent with Prithvi’s training strategy, but instead of reconstructing raw inputs, it learns to reconstruct the teacher’s MAE outputs—a process we term \textit{dual-MAE distillation} (Section~\ref{method1}). We then evaluated the compact variants across multiple downstream tasks by attaching task-specific heads and fine-tuning them on labelled datasets (Section~\ref{method2}). 

\begin{figure}[ht]
\centering
\newcommand{\cellw}{0.48\linewidth} 

\begin{tabular}{@{}c@{\hspace{0.02\linewidth}}c@{}}
\subcaptionbox{Cloud detection — tile-level classification\label{fig:cloud-classification}}
  {\includegraphics[width=\cellw]{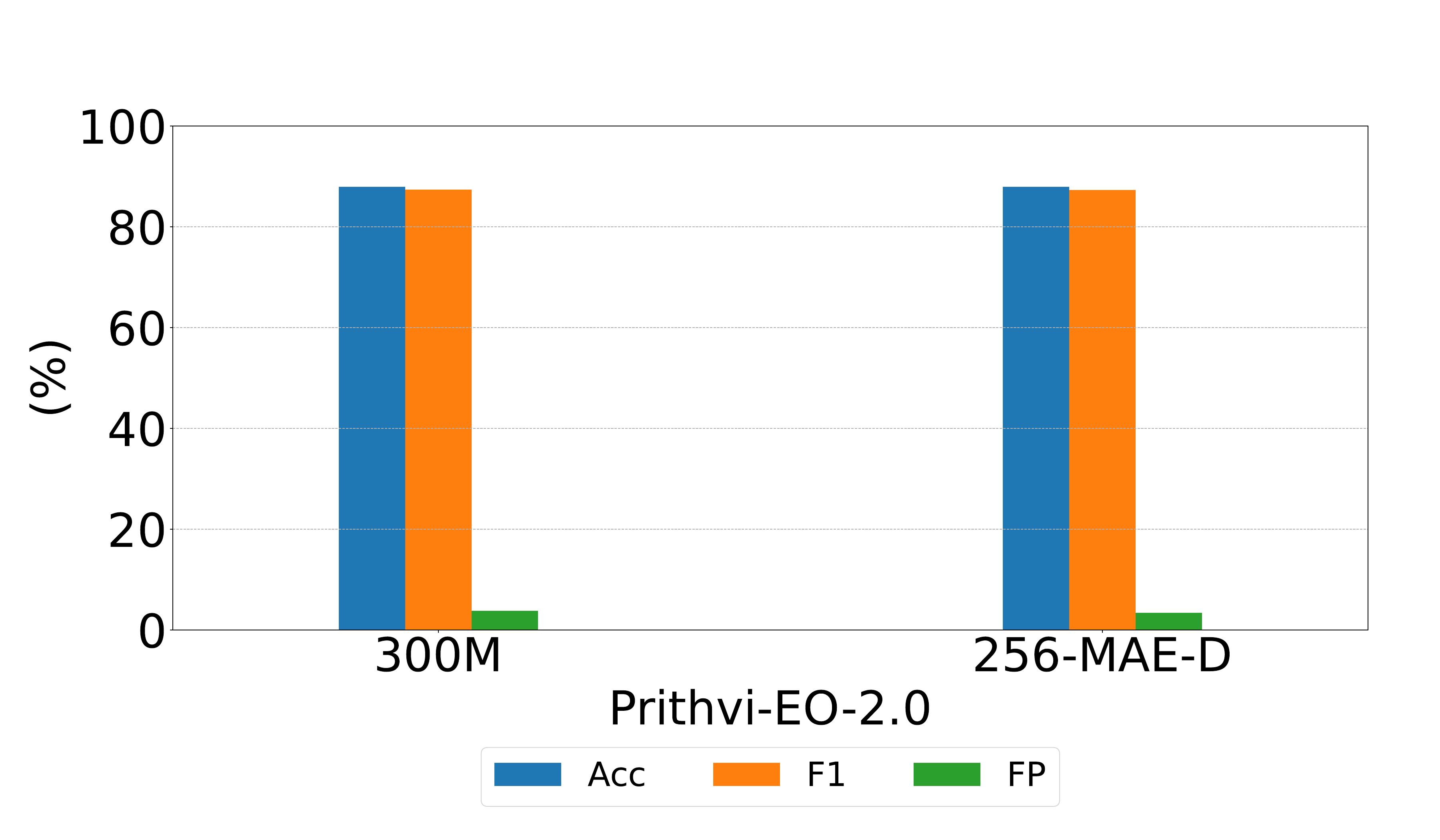}} &
\subcaptionbox{Cloud detection — segmentation\label{fig:cloud-segmentation}}
  {\includegraphics[width=\cellw]{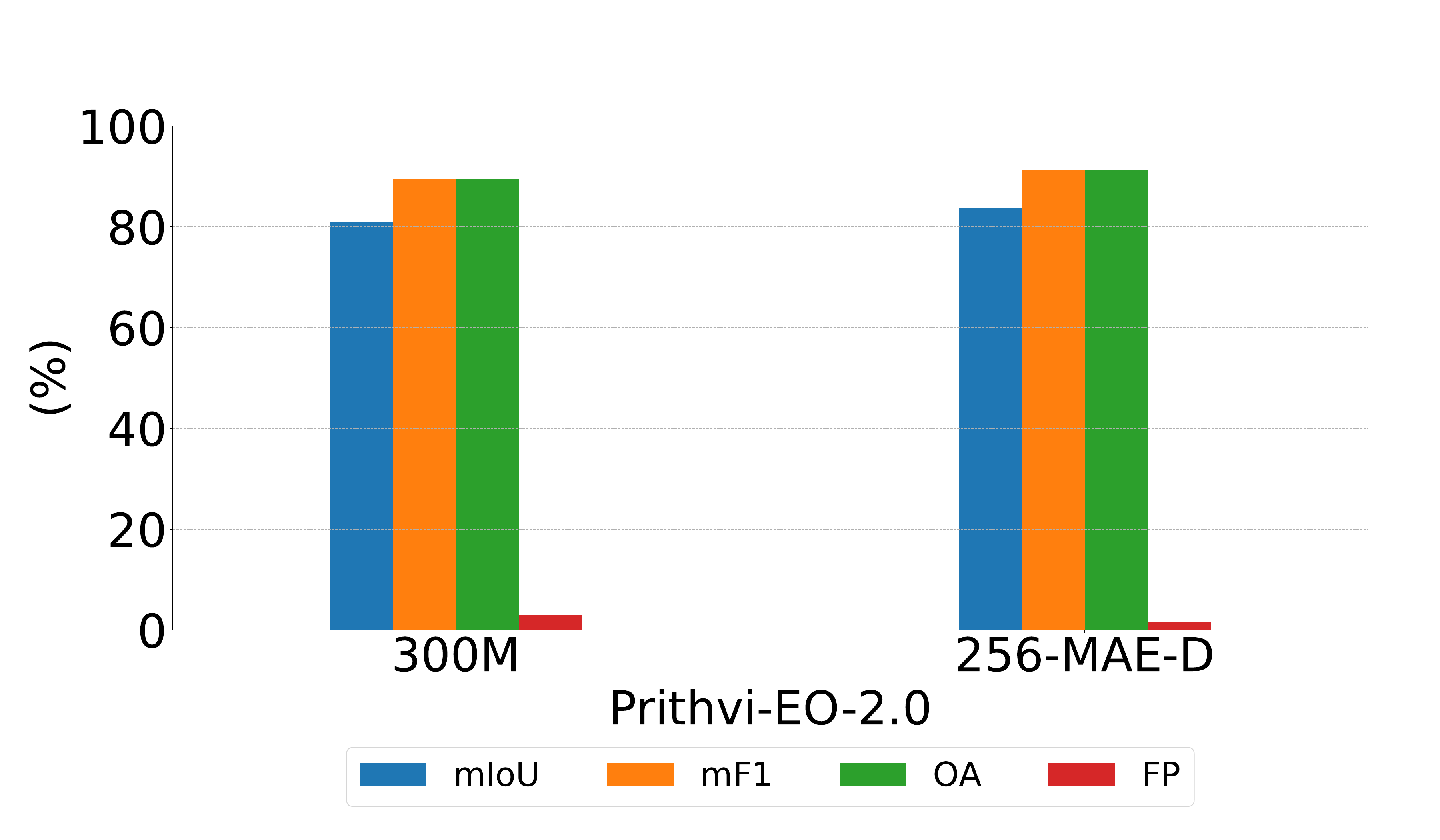}} \\[0.6em]

\subcaptionbox{Flood detection — segmentation\label{fig:flood-detection}}
  {\includegraphics[width=\cellw]{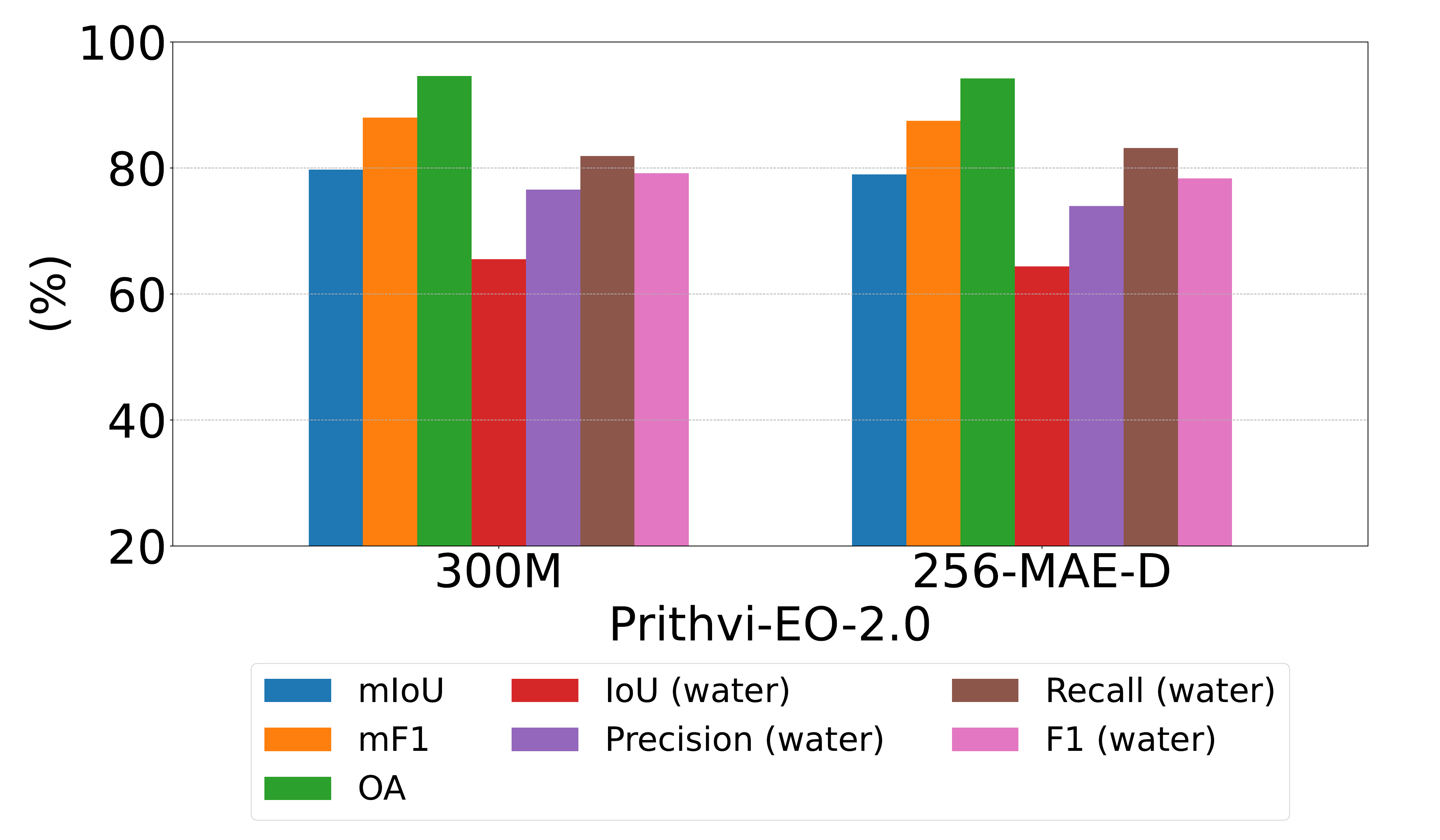}} &
\subcaptionbox{Landslide detection — segmentation\label{fig:landslide-detection}}
  {\includegraphics[width=\cellw]{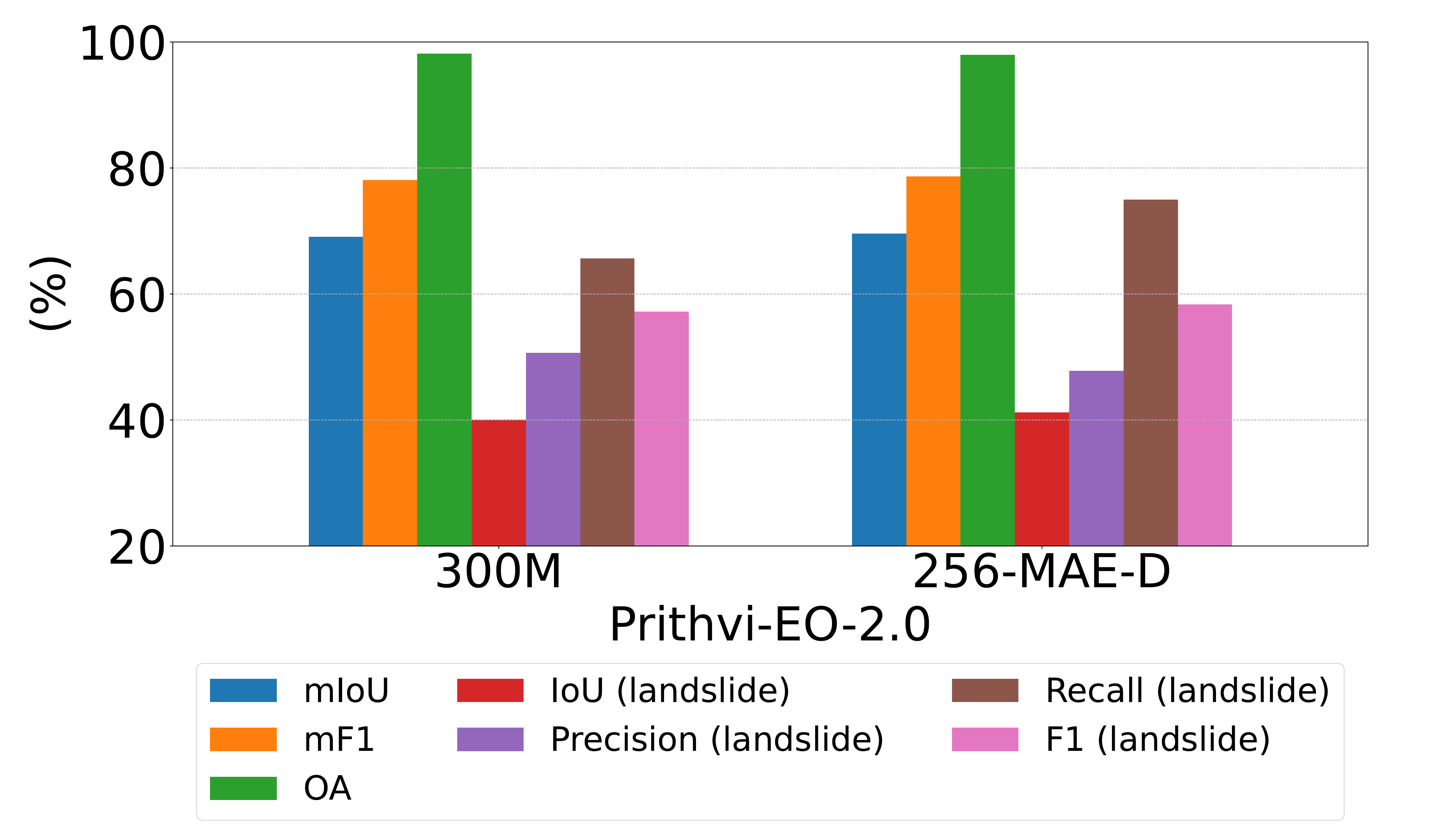}} \\[-0.7em]

\multicolumn{2}{c}{
\subcaptionbox{AGB estimation — regression\label{fig:agb}}
  {\includegraphics[width=\cellw]{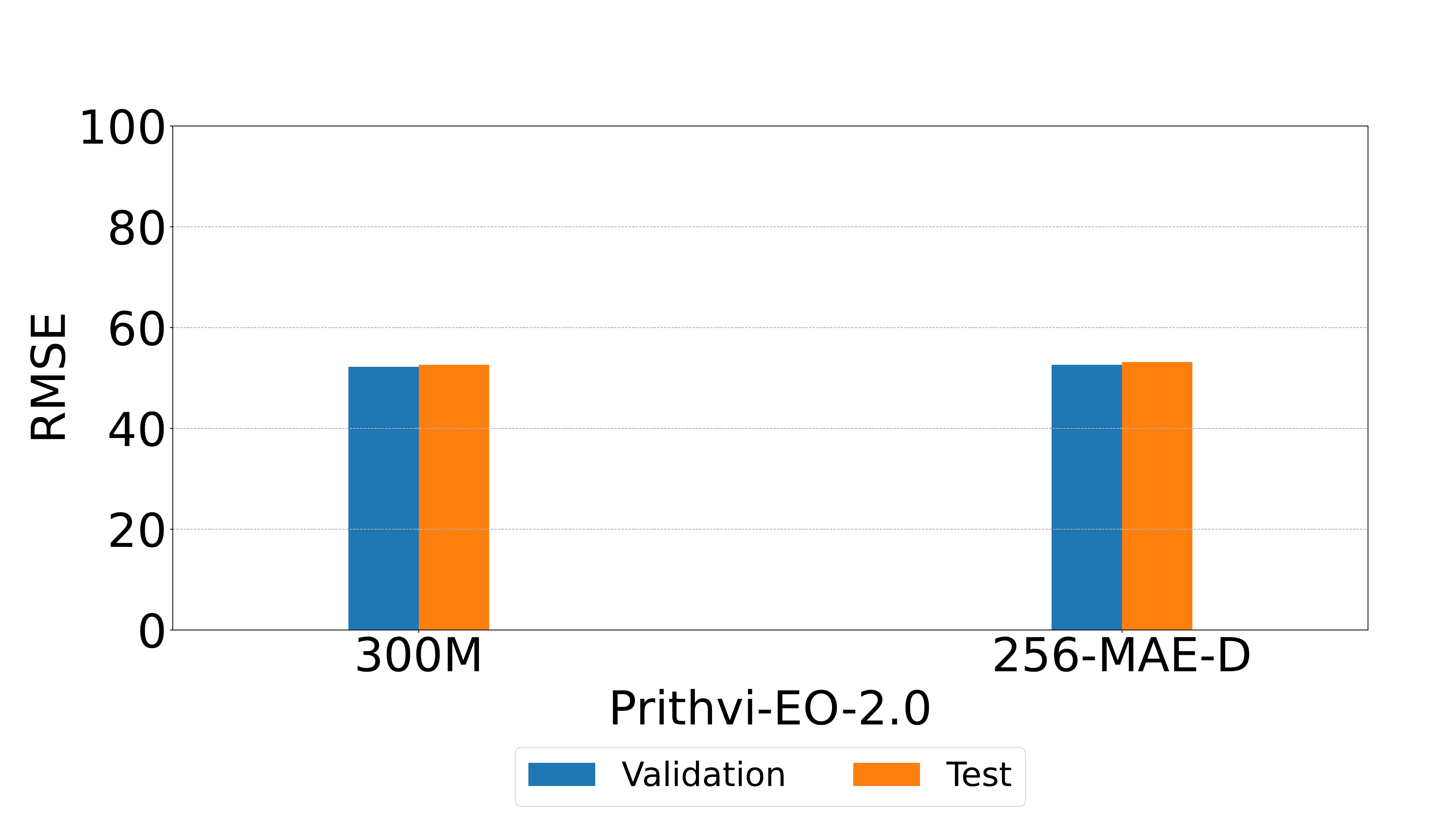}} 
} 


\end{tabular}

\caption{Performance comparison between the original Prithvi-300M~\cite{szwarcman2024prithvi} and its 16$\times$ smaller variant, 256-MAE-D, across five downstream tasks. Results span cloud detection~\cite{francis_alistair_2020_4172871} \textbf{(a, b)}, flood detection~\cite{bonafilia2020sen1floods11} \textbf{(c)}, landslide detection~\cite{ghorbanzadeh2022outcome,ghorbanzadeh2022landslide4sense} \textbf{(d)}, and AGB estimation~\cite{nascetti2023biomassters} \textbf{(e)}. 256-MAE-D delivers comparable performance across all tasks, showing that distillation preserves task generality while enabling onboard deployment.}
\label{fig:eo-task-results}
\end{figure}

Fig.~\ref{fig:eo-task-results} compares the performance of the 300M baseline and the distilled 256-MAE-D variant across five downstream tasks. Despite being $16\times$ smaller, 256-MAE-D delivers performance on par with the full model, with differences typically within run-to-run variability. For cloud detection, both tile-level classification and segmentation differed by less than one percentage point, with the compact model even achieving a lower false positive rate. Flood and landslide detection—both challenging due to class imbalance and scene variability—show similarly competitive results, indicating strong generalisation under limited ground-truth availability. For AGB estimation (Fig.~\ref{fig:agb}), RMSE remained within 3\% of the baseline. Collectively, these results show that aggressive compression through distillation preserves task generality while reducing the computational footprint to a level compatible with space-qualified hardware. 

Other compact Prithvi variants are shown in Supplementary Fig.~1. The MAE-only models (512-MAE, 256-MAE) reduce the patch embedding dimension to 512 and 256, and are trained solely with MAE, while 512-MAE-D applies dual-MAE distillation. These comparisons show that distillation improves robustness after compression, with 256-MAE-D offering the best balance of efficiency and performance for onboard use. We next evaluate how 256-MAE-D generalises to real mission data, particularly whether a domain gap exists. 

\subsection{Domain gap between source and target domain}\label{results-gap} 
\begin{figure}[ht]
\centering
\newcommand{\cellw}{0.48\linewidth} 
\begin{tabular}{@{}c@{\hspace{0.02\linewidth}}c@{}}
\subcaptionbox{Cloud detection — tile-level classification\label{fig:cloud-classification-domain-gap}}
{\includegraphics[width=\cellw]{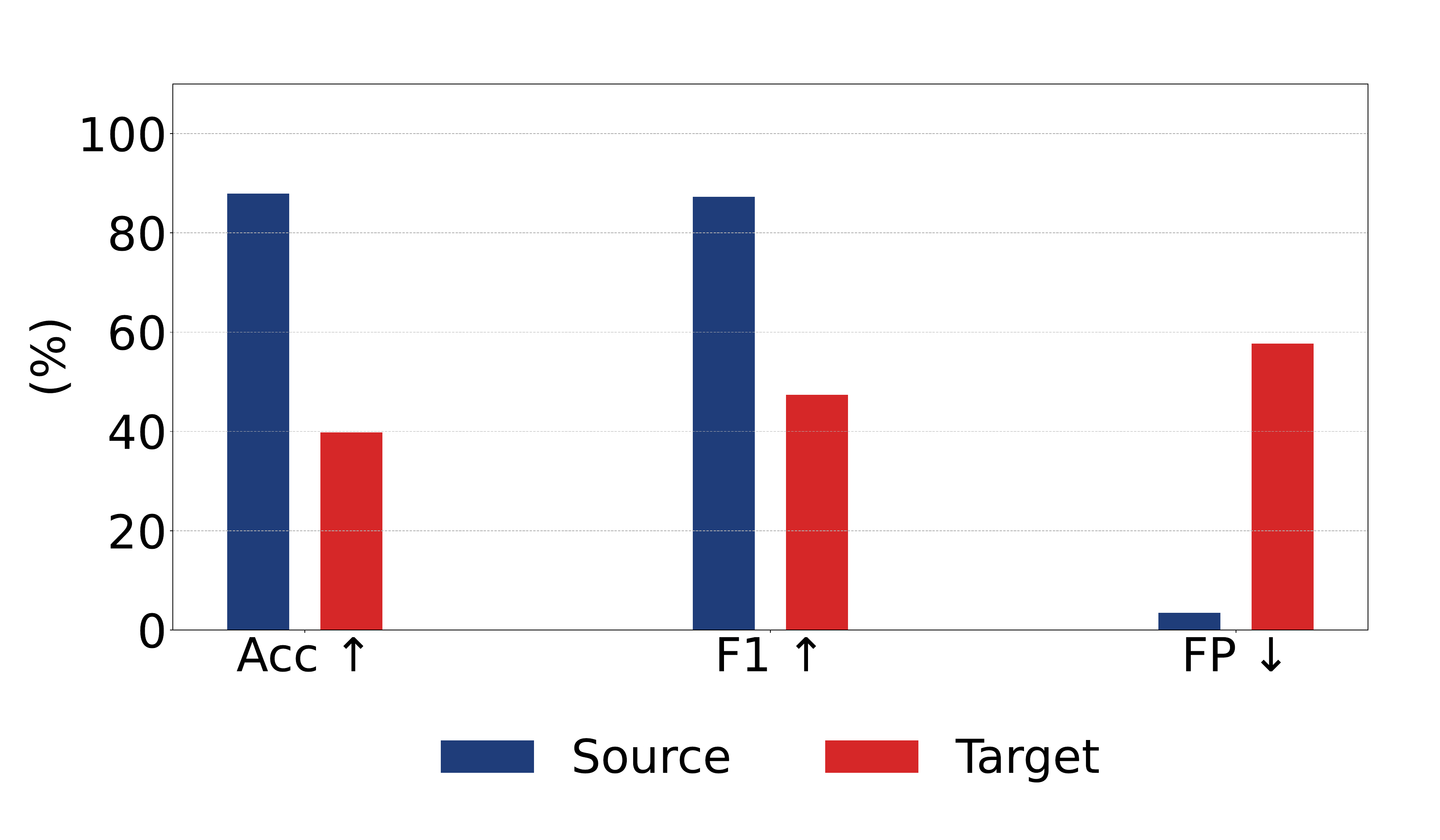}} &
\subcaptionbox{Cloud detection — segmentation\label{fig:cloud-segmentation-domain-gap}}
{\includegraphics[width=\cellw]{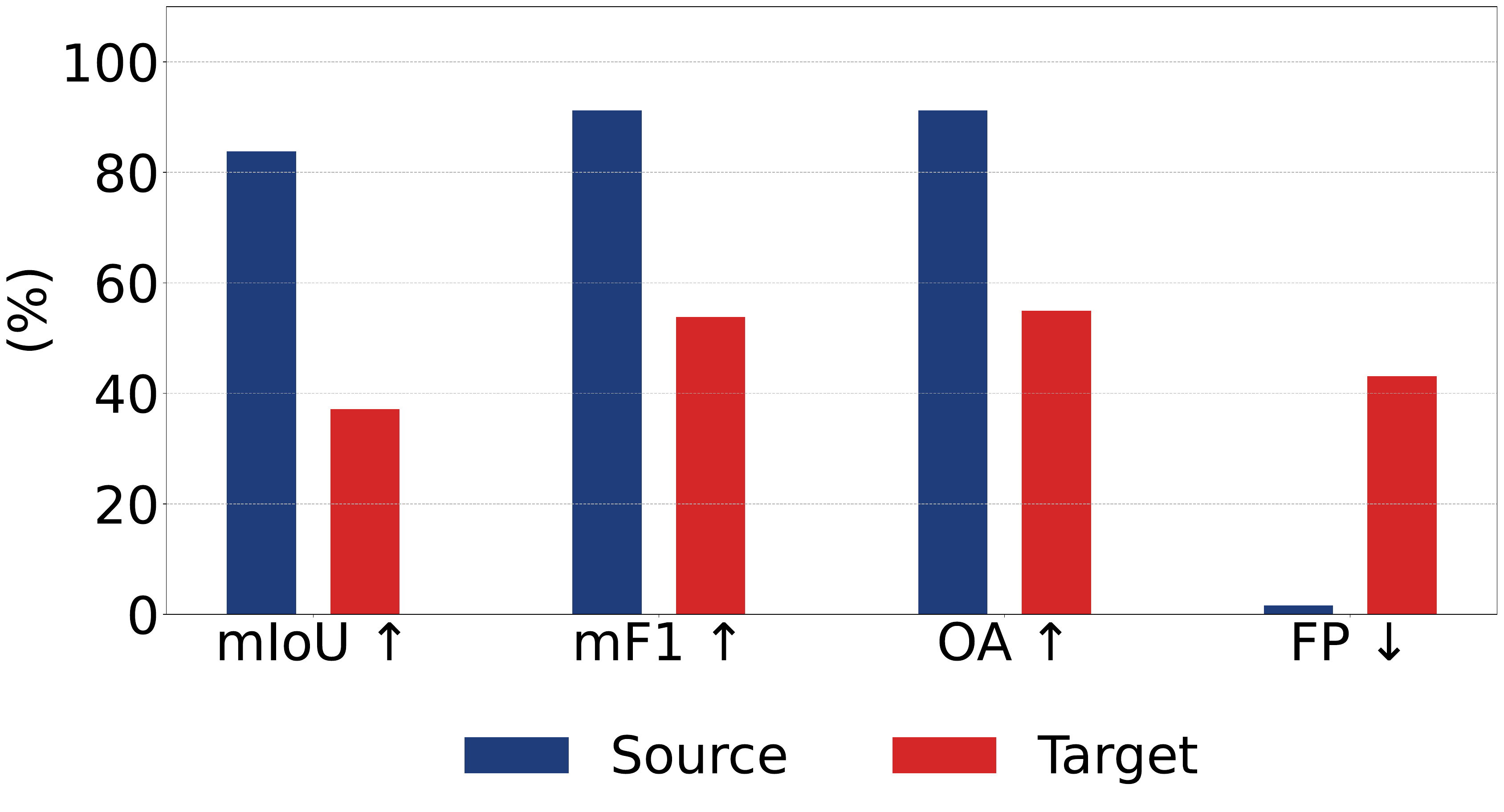}} 
\\[2.8em]

\multicolumn{2}{c}{
\subcaptionbox{Flood detection — segmentation\label{fig:flood-detection-domain-gap}}
{\includegraphics[width=\cellw]{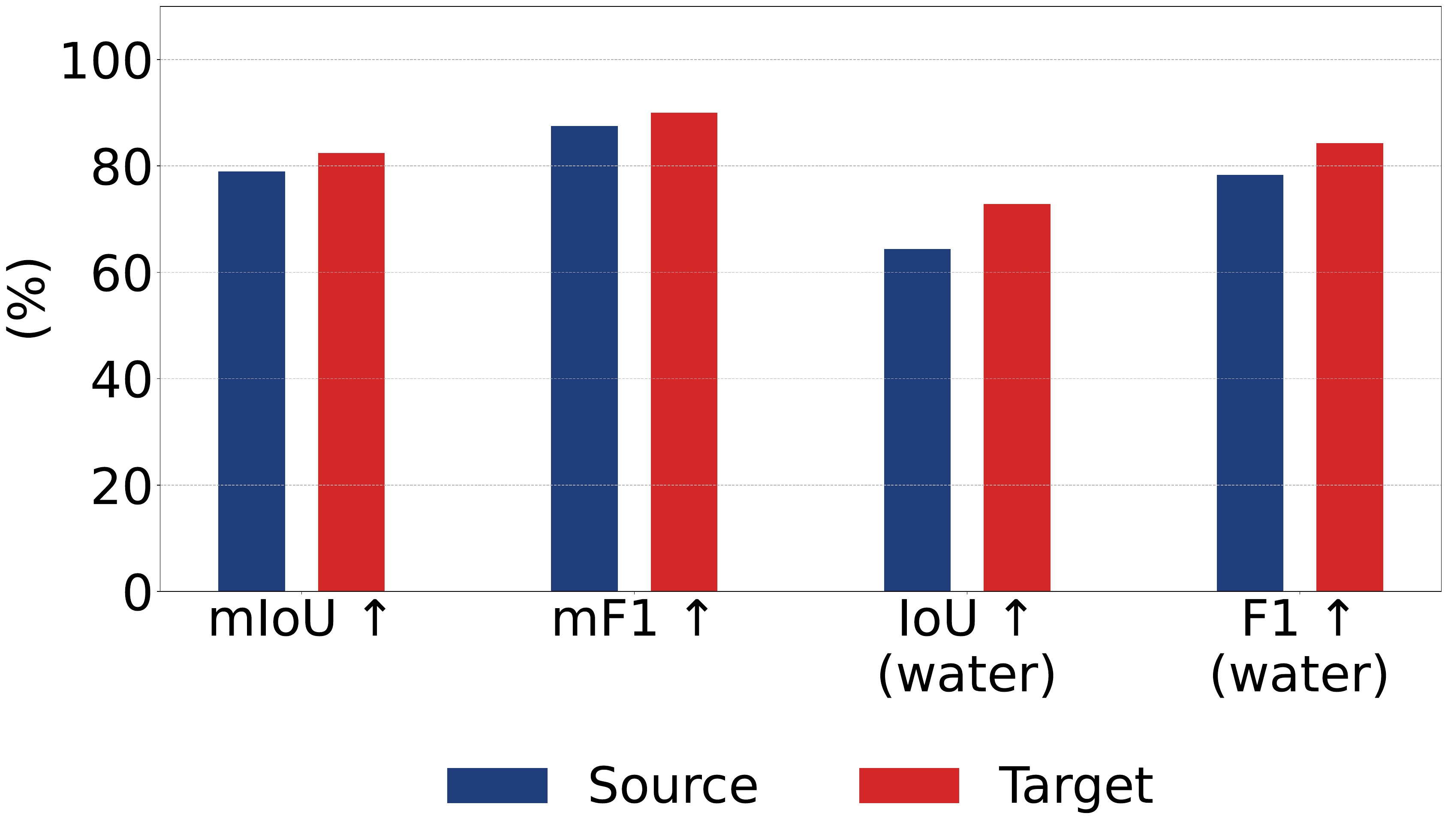}}  
} 
\end{tabular}
\caption{Performance of our GeoFM (256-MAE-D) on cloud and flood detection tasks, comparing held-out imagery from fine-tuning (source domain) with imagery from the Kanyini mission (target domain). Cloud detection \textbf{(a, b)} shows substantial declines across all metrics, whereas flood segmentation \textbf{(c)} exhibits modest gains.}
\label{fig:eo-domain-gap}
\end{figure}

After confirming that compression preserves benchmark performance, we examine how our GeoFM (256-MAE-D) generalises to real mission data, specifically whether a domain gap exists. To do this, we compared performance on data from the source domain (i.e., held-out imagery from the datasets used to fine-tune the task heads) with performance on data from the target domain (i.e., imagery captured from the Kanyini mission). We focused on cloud and flood detection as representative downstream tasks, since other tasks lacked suitable ground truth (Section~\ref{method3}). As shown in Fig.~\ref{fig:eo-domain-gap}, these comparisons reveal a clear domain gap, highlighting the challenges of transferring models across different satellite sensors.

Across the evaluated tasks, performance degradation in the target domain was substantial, particularly for cloud detection (Figs.~\ref{fig:cloud-classification-domain-gap}--\ref{fig:cloud-segmentation-domain-gap}). In the tile-level classification task, accuracy dropped by 48.2\% and F1 by 39.9\%, while the false-positive rate increased by 54.2\%. Similarly, for cloud segmentation, mIoU decreased by 46.7\%, mean F1 by 37.4\%, and overall accuracy by 36.2\%, with a 41.5\% rise in false positives. These large declines reflect a non-trivial domain gap between the source data (Sentinel-2) and the target data (Kanyini). A key driver is the spatial resolution mismatch: pretraining on Sentinel-2 (10–20 m) induces priors on textures and spatial frequencies that do not translate directly to Kanyini’s 75 m resolution, where features are mixed within single pixels and contextual information is altered. This is further compounded by radiometric and spectral differences. Variations in band response functions, calibration, and signal-to-noise ratios lead to systematically different reflectance values for the same surface material, making cross-sensor generalisation difficult without adaptation.

By contrast, flood detection (Fig.~\ref{fig:flood-detection-domain-gap}) exhibited an unexpected gain: mIoU improved by 3.4\%, mean F1 by 2.5\%, IoU for the water class by 8.5\%, and F1 for the water class by 6.0\%. This suggests that, in this case, the sensor characteristics of the Kanyini data may align more closely with the pretraining distribution, thereby reducing the severity of the domain gap. Taken together, these results confirm the presence of a non-trivial domain gap for cloud detection while also highlighting that its effects can be task-dependent, motivating the adaptation strategies explored in the following subsection. 

\subsection{Foundation model supports generalisation to the target domain and reduces training data needs}\label{results-adaptation}
\begin{figure}[H]
\centering
\begin{subfigure}[b]{0.90\linewidth}
    \includegraphics[width=\linewidth]{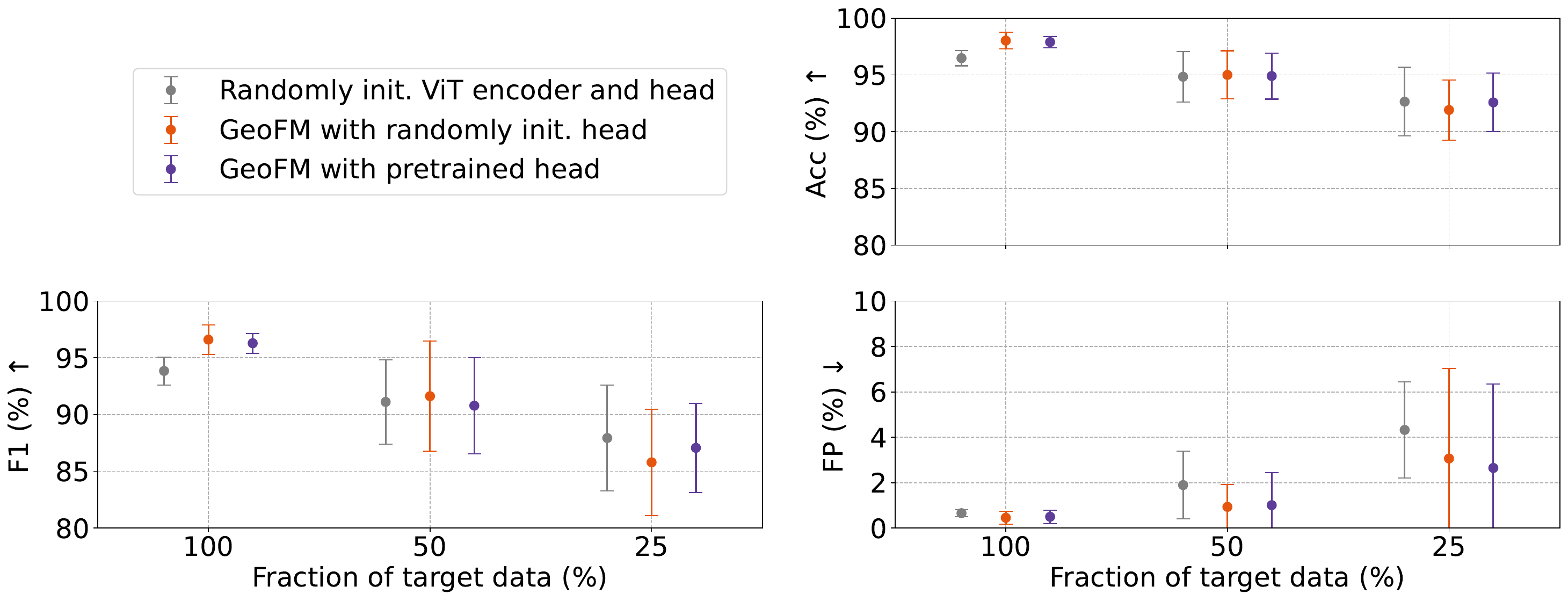}
    \caption{Cloud detection – tile-level classification. Baseline: Acc = 39.77\%, F1 = 57.70\%, FP = 47.41\%.}
    \label{fig:cloud-classification-kanyini}
\end{subfigure}
\begin{subfigure}[b]{0.90\linewidth}
    \includegraphics[width=\linewidth]{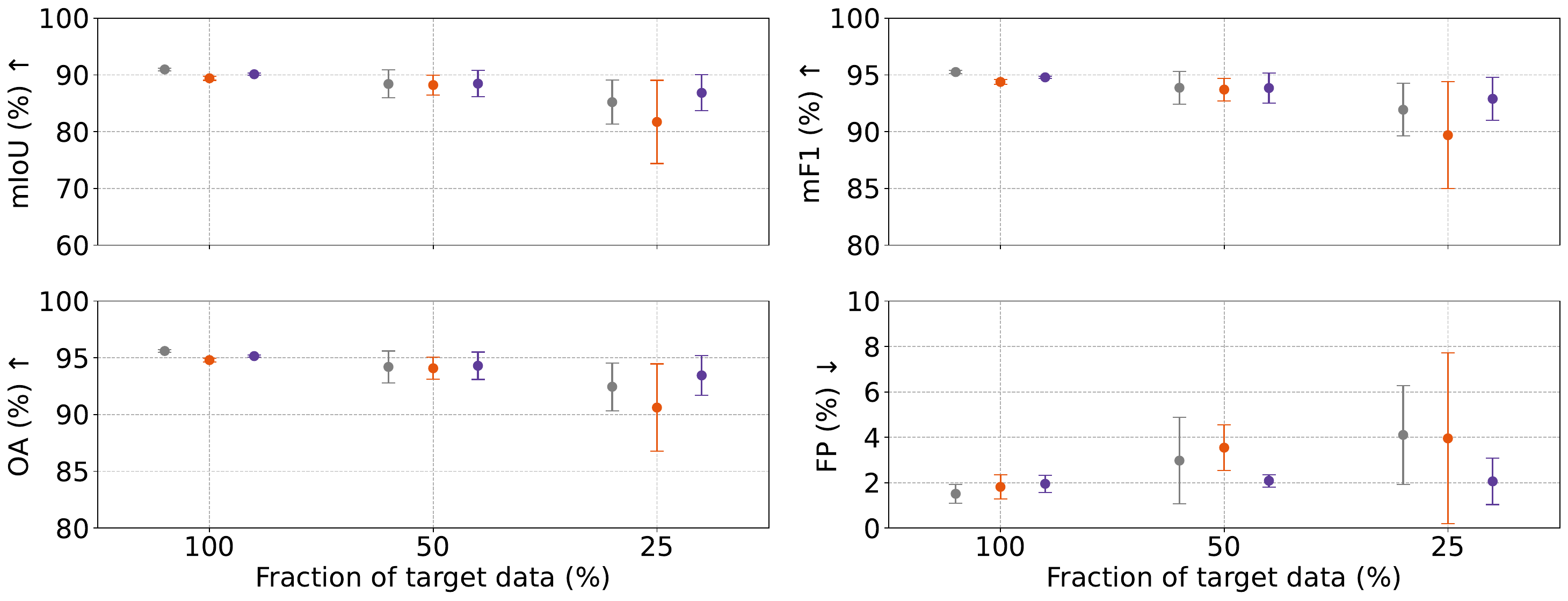}
    \caption{Cloud detection – segmentation. Baseline: mIoU = 37.17\%, mF1 = 53.83\%, OA = 54.97\%, FP = 43.11\%.}
    \label{fig:cloud-segmentation-kanyini}
\end{subfigure}
\begin{subfigure}[b]{0.90\linewidth}
    \includegraphics[width=\linewidth]{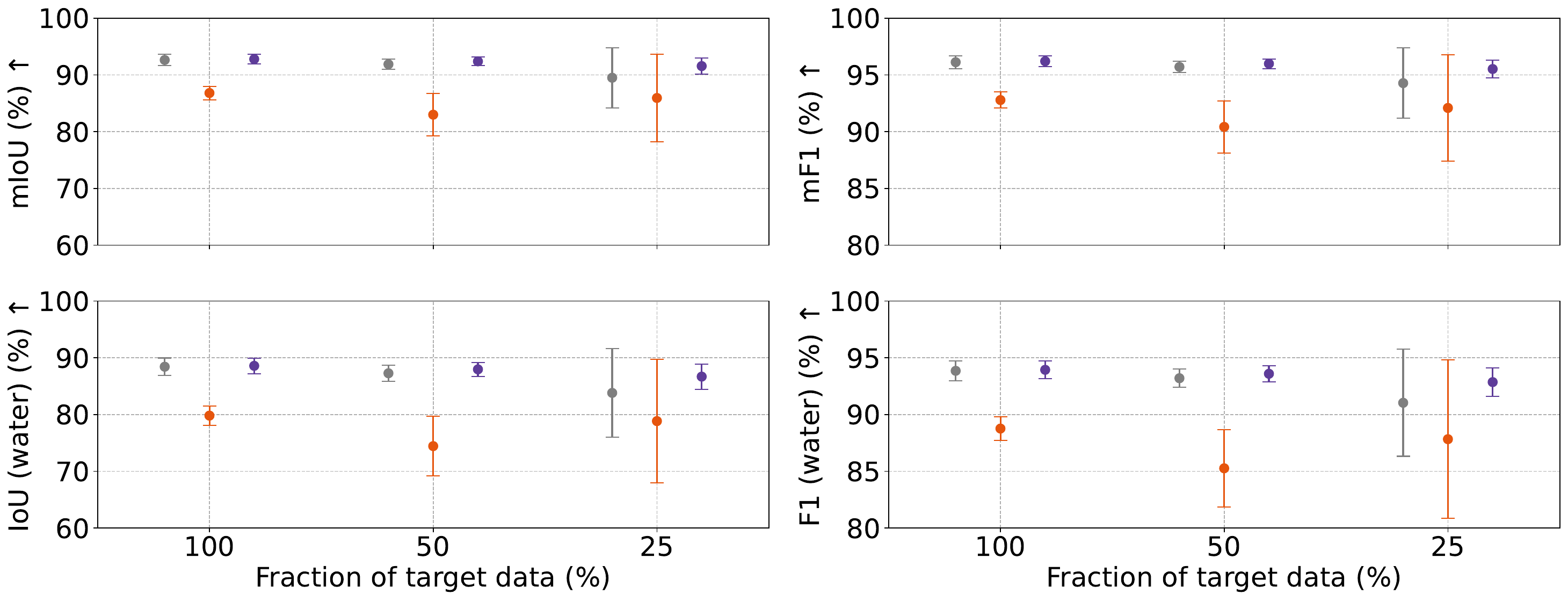}
    \caption{Flood detection – segmentation. Baseline: mIoU = 82.40\%, mF1 = 90.05\%, IoU (water) = 72.87\%, F1 (water) = 84.31\%.}
    \label{fig:flood-detection-kanyini}
\end{subfigure}

\caption{Effect of our GeoFM (256-MAE-D) on cloud and flood detection performance in the Kanyini domain under reduced labelled training data. Results are shown for tile-level cloud classification \textbf{(a)}, cloud segmentation \textbf{(b)}, and flood segmentation \textbf{(c)}, with error bars denoting standard deviation across five random seeds. Baseline values in each subfigure indicate performance \emph{before domain adaptation}, i.e., direct transfer from the source to the Kanyini domain.}

\label{fig:training-data-results}
\end{figure}

Given the observed domain gap, we evaluate whether domain adaptation enables our GeoFM (256-MAE-D) to generalise to the Kanyini domain and whether it improves data efficiency (Section~\ref{method3}). We train three configurations for cloud and flood detection (Fig.~\ref{fig:training-data-results}): (1) a randomly initialised ViT encoder with a randomly initialised head; (2) our GeoFM with a randomly initialised head; and (3) our GeoFM with a head pretrained on source-domain data. In (2) and (3), the GeoFM encoder is frozen and only the head is retrained. Each configuration is trained with 100\%, 50\%, and 25\% of the labelled data, with results averaged over five seeds. This setup allowed us to evaluate both how well our GeoFM transfers to the Kanyini domain and its impact on data efficiency. 

Pretraining provided consistent benefits across tasks, with the largest gains appearing as labelled data decreased. In cloud classification and segmentation (Figs.~\ref{fig:cloud-classification-kanyini}--\ref{fig:cloud-segmentation-kanyini}), both GeoFM configurations outperformed the randomly initialised baseline, and the \textit{GeoFM with pretrained head} achieved the best overall accuracy, F1, mIoU, and false-positive rates. Flood detection (Fig.~\ref{fig:flood-detection-kanyini}) showed the same trend, with one nuance: the \textit{GeoFM with randomly init.\ head} underperformed the baseline at 100\% data, indicating that an untrained head may not fully leverage the GeoFM’s representations. The pretrained head closed this gap and delivered stronger improvements in low-data settings, especially for water-class IoU and F1. Overall, task-head pretraining, while not always essential, helps align decision boundaries and stabilises transfer when task- or domain-specific features differ. 

Most importantly, these findings address the guiding questions in Section~\ref{method3}. Frist, our GeoFM substantially reduces the amount of labelled data required, performing strongly even with only 25\% of labels. Second, the GeoFM can be deployed without adaptation (i.e., kept frozen), as shown in cloud detection. However, this may be insufficient, as shown in flood detection, particularly when the head is not pretrained. Third, task-head pretraining is not always essential, but it consistently boosts performances in low-data regimes, and is critical when the GeoFM underperforms as shown in flood detection when a randomly initialised head is used. Overall, both GeoFM configurations benefit from domain adaptation, with the pretrained head offering the most reliable pathway for adapting to the Kanyini domain. We next evaluate whether our GeoFM can also be executed on flight-representative hardware.

\subsection{Foundation model executes on flight-representative hardware}\label{results-hardware}
With the domain gap mitigated, we next validate whether our GeoFM (256-MAE-D) can be executed accurately and efficiently on flight-representative hardware from two EO missions (Section~\ref{method4}). For Kanyini, we evaluated the model on a hardware emulator (Supplementary Fig. 12a) featuring the Ubotica CogniSAT-XE1 and then on the HyperScout-2 engineering model (Supplementary Fig. 12b) featuring the Eyes of Things (EOT) board. Both platforms incorporate the Myriad-2 accelerator, and in the emulator setup, the XE1 was paired with an embedded system matching the CPU and RAM of the HyperScout-2 data-handling unit, ensuring identical computational constraints and a shared software stack. In both environments, the GeoFM executed directly on the Myriad-2. For IMAGIN-e, validation was performed on a Raspberry Pi~4 (RPi~4) Model~B and then on the IMAGIN-e engineering model (Supplementary Fig. 13). The RPi~4 matches the CPU architecture of the mission module but with fewer cores and less RAM; in both cases, execution ran on the CPU as no hardware accelerator is available.

Table~\ref{tab:kanyini-performance} shows that reducing model weights from FP32 to FP16, as required by the Myriad-2, has negligible impact on performance across all three EO tasks. Cloud classification accuracy changed only from 97.22\% (FP32) to 97.10\% (FP16), and cloud segmentation mIoU from 90.14\% to 90.12\%. For flood detection, mIoU differed by just 0.11 percentage points. Across all metrics, FP32--FP16 deltas were $\leq$\,0.21 percentage points. Beyond preserving performance, FP16 halves the storage footprint, reducing uplink volume and transmission time for model updates while simplifying deployment on resource-constrained hardware. Qualitative outputs from the HyperScout-2 engineering model further corroborate these results, with predicted masks closely matching ground-truth labels for both cloud (Supplementary Fig.~2) and flood detection (Supplementary Fig.~3).

Importantly, the hardware emulator and engineering model, both running FP16 on the Myriad-2, produced indistinguishable results, confirming the emulator as a reliable proxy for flight hardware. Likewise, the RPi~4 matched the workstation baseline when executing in FP32 on CPU, and the IMAGIN-e engineering model produced identical outputs under the same conditions.

Table~\ref{tab:kanyini-efficiency} summarises efficiency and resource use when running inference on a single Kanyini data cube. Inference performance was highly consistent across both environments: per-tile runtimes were effectively identical (e.g., 5.36\,s vs.\ 5.35\,s for cloud classification), and total cube runtimes differed by at most 25\,s. Memory usage remained stable at $\sim$630--640\,MB. The main difference was power consumption: both peak and average power were higher on the engineering model (e.g., 7.26\,W vs.\ 5.76\,W peak), leading to greater overall energy use (0.98--1.07\,Wh vs.\ 0.72--0.79\,Wh). Thus, the emulator offers a faster, more accessible prototyping environment, while the engineering model more accurately reflects the power envelope of space-qualified hardware. Efficiency profiling was not performed on the RPi~4, which served only to confirm CPU-based feasibility for IMAGIN-e, as it is not an exact hardware match to the mission module.

Together, these results show that our GeoFM executes reliably on space-qualified hardware without loss of performance, completing the progression from model compression and domain adaptation to deployment under flight-representative conditions. We next extend this validation to an on-orbit demonstration using the IMAGIN-e mission.

\begin{table*}[ht]
\centering
\small
\caption{Performance across three downstream tasks on the Kanyini test sets. 
All metrics are reported as percentages. Results compare FP32 and FP16 precision across four environments: workstation (GPU, FP32), hardware emulator (XE1, Myriad-2, FP16, Kanyini), HyperScout-2 engineering model (EOT, Myriad-2, FP16, Kanyini), Raspberry Pi 4 (CPU, FP32, IMAGIN-e), and IMAGIN-e engineering model (CPU, FP32, IMAGIN-e).}

\label{tab:kanyini-performance}

\begin{subtable}{\linewidth}
\centering
\small
\label{tab:cloud_classification_combined}
\begin{tabular}{l c c c c c}
\toprule
\textbf{Environment} & \textbf{\shortstack{Hardware \\ Accelerator}} & \textbf{\shortstack{FP \\ Format}}
& \textbf{Acc} & \textbf{FP} & \textbf{F1} \\
\midrule
\rowcolor{gray!15}\multicolumn{6}{c}{\textbf{Cloud detection  tile-based classification (792 tiles)}}\\
Workstation       & GPU            & FP32 & 97.22 & 0.76 & 95.22 \\
Hardware emulator & XE1 (Myriad-2) & FP16 & 97.10 & 0.88 & 95.01 \\
HyperScout-2 EM     & EOT (Myriad-2) & FP16 & 97.10 & 0.88 & 95.01 \\
Raspberry Pi 4    & CPU            & FP32 & 97.22 & 0.76 & 95.22 \\
IMAGIN-e EM       & CPU            & FP32 & 97.22 & 0.76 & 95.22 \\
\bottomrule
\end{tabular}
\end{subtable}

\vspace{0.75em}

\begin{subtable}{\linewidth}
\centering
\small
\label{tab:cloud_segmentation_combined}
\begin{tabular}{l c c c c c c}
\toprule
\textbf{Environment} & \textbf{\shortstack{Hardware \\ Accelerator}} & \textbf{\shortstack{FP \\ Format}}
& \textbf{mIoU} & \textbf{mF1} & \textbf{OA} & \textbf{FP} \\
\midrule
\rowcolor{gray!15}\multicolumn{7}{c}{\textbf{Cloud detection - segmentation (792 tiles)}}\\
Workstation       & GPU            & FP32 & 90.14 & 94.80 & 95.18 & 1.50 \\
Hardware emulator & XE1 (Myriad-2) & FP16 & 90.12 & 94.79 & 95.17 & 1.52 \\
HyperScout-2 EM     & EOT (Myriad-2) & FP16 & 90.12 & 94.79 & 95.17 & 1.52 \\
Raspberry Pi 4    & CPU            & FP32 & 90.14 & 94.80 & 95.18 & 1.50 \\
IMAGIN-e EM       & CPU            & FP32 & 90.14 & 94.80 & 95.18 & 1.50 \\
\bottomrule
\end{tabular}
\end{subtable}

\vspace{0.75em}

\begin{subtable}{\linewidth}
\centering
\small
\label{tab:flood_segmentation_combined}
\begin{tabular}{l c c c c c c}
\toprule
\textbf{Environment} & \textbf{\shortstack{Hardware \\ Accelerator}} & \textbf{\shortstack{FP \\ Format}}
& \textbf{mIoU} & \textbf{mF1} & \multicolumn{2}{c}{\textbf{Water}} \\
\cmidrule(lr){6-7}
 &  &  &  &  & \textbf{IoU} & \textbf{F1} \\
\midrule
\rowcolor{gray!15}\multicolumn{7}{c}{\textbf{Flood detection - segmentation (880 tiles)}}\\
Workstation       & GPU            & FP32 & 91.45 & 95.46 & 86.49 & 92.76 \\
Hardware emulator & XE1 (Myriad-2) & FP16 & 91.56 & 95.53 & 86.64 & 92.84 \\
HyperScout-2 EM     & EOT (Myriad-2) & FP16 & 91.56 & 95.53 & 86.64 & 92.84 \\
Raspberry Pi 4    & CPU            & FP32 & 91.45 & 95.46 & 86.49 & 92.76 \\
IMAGIN-e EM       & CPU            & FP32 & 91.45 & 95.46 & 86.49 & 92.76 \\
\bottomrule
\end{tabular}
\end{subtable}
\end{table*}

\begin{table*}[ht]
\centering
\small
\caption{Efficiency and resource utilisation per Kanyini data cube (88 tiles). 
Metrics include per-tile inference time, total runtime, peak memory, peak and average power, and energy consumption. Results are reported for the hardware emulator and HyperScout-2 engineering model, both representing the Kanyini mission hardware. Profiling was not performed on the RPi~4, as it was used only to verify feasibility of execution on a CPU platform similar to the IMAGIN-e compute module.}

\label{tab:kanyini-efficiency}
\setlength{\tabcolsep}{6pt}
\begin{tabular}{l c c c c c c}
\toprule
\textbf{Environment} 
& \textbf{\shortstack{Inf time \\ (per tile)}} 
& \textbf{Runtime} 
& \textbf{\shortstack{Peak \\ Memory}} 
& \textbf{\shortstack{Peak \\ Power}} 
& \textbf{\shortstack{Avg \\ Power}} 
& \textbf{Energy} \\
& (s) & (s) & (MB) & (W) & (W) & (Wh) \\
\midrule

\rowcolor{gray!15}\multicolumn{7}{c}{\textbf{Cloud detection - tile-based classification}} \\
Hardware emulator & 5.36 & 526.41 & 636.25 & 5.76 & 4.95 & 0.72 \\
HyperScout-2 EM & 5.35 & 551.66 & 633.58 & \textbf{7.26} & \textbf{6.41} & 0.98 \\
\midrule

\rowcolor{gray!15}\multicolumn{7}{c}{\textbf{Cloud detection - segmentation}} \\
Hardware emulator & 5.50 & 553.21 & 633.45 & 5.77 & 4.94 & 0.76 \\
HyperScout-2 EM & 5.49 & 575.34 & 636.09 & \textbf{7.16} & \textbf{6.39} & 1.02 \\
\midrule

\rowcolor{gray!15}\multicolumn{7}{c}{\textbf{Flood detection - segmentation}} \\
Hardware emulator & 5.68 & 573.42 & 638.18 & 6.20 & 4.97 & 0.79 \\
HyperScout-2 EM & 5.67 & 597.43 & 631.77 & \textbf{7.21} & \textbf{6.45} & 1.07 \\
\bottomrule
\end{tabular}
\end{table*}


\subsection{Foundation model successfully demonstrated on an EO space platform}\label{results-deployment}
Having confirmed that our GeoFM (256-MAE-D) executes consistently and without error on flight-representative hardware, we next validated its behaviour under true orbital conditions. To this end, the model was uplinked to IMAGIN-e, an EO payload aboard the ISS, and executed in orbit under ground command to verify that a compact GeoFM can operate as intended on a resource-constrained space platform. A curated test set representative of typical mission scenes was also uplinked and used to perform inference for cloud and flood detection directly on-board (see Section~\ref{method5} for details of the on-orbit procedure). 

During initial runs, the GeoFM’s intensive use of processing resources caused the on-board compute module to approach its thermal limits, triggering automatic shutdowns. This behaviour was not observed during laboratory validation, likely due to differences in the thermal environment and power protection thresholds between the engineering and flight models. To ensure stable operation within safe thermal margins, the number of active CPU cores was reduced. With this configuration, all experiments completed successfully and the system remained thermally stable throughout execution. Task-level performance metrics obtained on IMAGIN-e exactly matched those from the RPi~4 and engineering model, confirming numerical equivalence between laboratory validation and on-orbit execution. 

Following this validation, the next planned step was to execute the GeoFM on newly acquired imagery; however, this was not possible due to a power fault in the imaging system during the test window. Nonetheless, these experiments represent the first successful demonstration of a GeoFM deployed on an EO payload in space, marking an important step towards operational adoption in future missions.

\section{Discussion}\label{discussion} 

In this study, we demonstrate the feasibility of deploying a GeoFM on resource-constrained space platforms, marking a shift in how AI can support EO missions. By compressing a large-scale pretrained transformer-based model, adapting it to new domains, and validating its execution under genuine orbital conditions, we address four key challenges for on-orbit AI: the prohibitive size of modern foundation models, the strict computational and power constraints of satellites, the distribution shifts between training and operational data (domain gap), and the scarcity of labelled data for new missions. Together, these advances demonstrate that foundation models can be adapted for spaceborne deployment, taking an important step beyond laboratory validation toward operational use in EO missions. 

Despite these advances, several limitations should be acknowledged. On IMAGIN-e, validation was constrained by a power fault in the imaging system, requiring the use of a curated test set rather than live acquisitions. On Kanyini, compatibility with our GeoFM required reducing its rich hyperspectral data to four bands---RGB and near-infrared (NIR)---leaving much of the information from its 50 visible and near-infrared (VNIR) bands untapped. Our domain adaptation experiments were limited to cloud and flood detection tasks, as other applications lacked suitable ground truth, preventing evaluation across a broader range of downstream tasks such as landslide detection and AGB estimation. Finally, model deployment remains strongly dependent on the onboard runtime environment, which constrains how easily the deployment pipeline can be applied across platforms. On Kanyini, execution on the Myriad-2 VPU required conversion into a supported model format, along with adaptations to FP16 precision and the restricted operator set. By contrast, IMAGIN-e, which lacked a dedicated accelerator, allowed models to be run in a variety of formats provided the necessary software and libraries were uploaded. This variability shows that different hardware imposes different toolchain requirements, and while our framework is compatible with both VPUs and CPU-based platforms, extending deployment to future accelerators will require tailored optimisation pathways. In all cases, uplink bandwidth constraints must also be respected, further shaping the choice of model formats and software environments. Together, these factors define the current scope of our findings and highlight the need for further validation. 

Beyond the technical contributions, these findings have important implications for the role of AI in future EO missions. Task-specific CNNs have already shown that satellites can perform a variety of EO tasks~\cite{esa-phisat1, esa-phisat2, mateo2023orbit, ruzicka2023fast, schwarz2025early, rijlaarsdam2024next}. This reduces reliance on high-volume downlinks and enables rapid delivery of mission-critical insights, a capability especially valuable for disaster response~\cite{tellman2021satellite, ainscoe2025earthquake}, wildfire monitoring~\cite{lu2024onboard}, flood mapping~\cite{tellman2021satellite}, and maritime domain awareness~\cite{goudemant2024onboard}. However, CNN-based approaches typically require a separate model to be trained and uplinked for each new application, increasing bandwidth costs and operational overhead associated with repeated data collection, training, and model management. By contrast, GeoFMs reduce these burdens by reusing a single pretrained backbone: new tasks can be supported with lightweight task heads that require fewer labelled samples and minimal uplink bandwidth, streamlining both data preparation and mission operations. This efficiency enables satellites to expand their capabilities over time more readily than with task-specific models, offering a more scalable pathway for evolving missions. In line with recent calls for responsible and resource-efficient GeoFMs~\cite{natmi2025_responsible_geofm}, this work demonstrates a practical route for tailoring compact foundation models to space-qualified hardware, advancing both the technical feasibility and sustainable use of AI in operational EO monitoring. 

Several promising directions follow from this work. A key next step is to conduct live-data demonstrations on future missions to validate the reliability of onboard inference under genuine operational conditions~\cite{giuffrida2022phisat1, rijlaarsdam2024next}. Such tests will provide stronger evidence of performance at scale, across sustained operations and diverse observation conditions, and under real-world constraints such as illumination changes, communication dropouts, and thermal fluctuations.

Another priority is to expand evaluations beyond cloud and flood detection. Applications involving hyperspectral imaging for vegetation stress, crop and species discrimination, and water quality monitoring, or synthetic aperture radar (SAR) for all-weather, day–night coverage, will broader assessments of compact GeoFM adaptability. Progress in space-qualified accelerators, including emerging low-power GPUs, will expand the range of deployable models and multimodal pipelines, though advances in compression and adaptation will remain essential to balance efficiency and performance.

Automated design and improved data generation also present opportunities. Neural architecture search (NAS)~\cite{ren2021comprehensive, del2025optimizing} can produce architectures tailored to onboard constraints while high-fidelity simulators~\cite{longepe2024simulation} can generate mission-specific datasets to reduce reliance on costly annotations and accelerate mission readiness. Continual model adaptation---potentially even updating models directly in orbit~\cite{du2024domain}---is a natural extension, leveraging the broad representations learned by foundation models.

Finally, robustness and scalability remain open challenges. Spaceborne systems must be resilient to digital and physical perturbations to ensure trustworthy deployment~\cite{du2022adversarial}, while future constellations of satellites will require autonomous pipelines capable of coordinating, prioritising, and interpreting data streams in real time. Together, these directions outline a roadmap for fully autonomous, foundation model–driven EO systems acting as intelligent agents in orbit.

\section{Methods}\label{methods}

\subsection{Mission context}\label{method0}
\subsubsection*{Kanyini mission}
Kanyini is a 6U CubeSat developed through a South Australian Government initiative to strengthen sovereign EO capability~\cite{sasic_sasat1_space_services}. The mission was delivered in partnership with SmartSat CRC~\cite{smartsatcrc_homepage} (mission lead), Inovor Technologies~\cite{inovor_homepage} (spacecraft prime using the Apogee bus), and Myriota~\cite{myriota_homepage} (developer of the low-power IoT communications payload). The satellite carries two primary payloads: (i) Cosine's~\cite{cosine_homepage} HyperScout-2 hyperspectral imager, providing EO imaging capability, and (ii) Myriota’s IoT Space Services device for low-power data relay. HyperScout-2 integrates a dedicated Myriad-2 VPU, enabling on-orbit execution of compact ML models. Details of the HyperScout-2 instrument and compute subsystem are provided in Section~\ref{method4}. \\

Kanyini was launched on 17~August~2024 aboard SpaceX’s Falcon~9 as part of the Transporter-11 rideshare mission. After deployment into sun-synchronous orbit, the spacecraft completed standard bus and payload commissioning, during which the HyperScout-2 payload returned its first hyperspectral images and the Myriota payload demonstrated on-orbit IoT data relay. In mid-2025, the mission encountered an unexpected critical technical anomaly that compromised the satellite’s payload interface computer, preventing further communication with both the hyperspectral imager and the IoT payload. All other spacecraft subsystems remain fully operational, allowing continued monitoring through telemetry, tracking, and control channels. This anomaly does not affect the results presented in this work, which rely on the HyperScout-2 engineering model and a representative on-orbit compute environment.

\subsubsection*{IMAGIN-e payload}
IMAGIN-e is an ISS-hosted payload jointly developed by Thales Alenia Space~\cite{thales_alenia_space_homepage} and Microsoft~\cite{microsoft_homepage} to demonstrate AI-driven EO processing in orbit. The payload comprises a compact ARM-based compute module without hardware acceleration. Two imaging sensors are integrated into the system: an RGB camera (operational but currently unfocused) and Cosine's HyperScout-M hyperspectral imager, which remains non-operational due to a power issue. Applications are deployed on a K3S Kubernetes cluster, require Docker images built for the \texttt{aarch64} architecture, and execute entirely offline with no internet connectivity. A flight-representative engineering model of the compute module is available on the ground and is used for all experiments reported in this work. Details of the compute architecture and onboard software environment are provided in Section~\ref{method4}. \\

The IMAGIN-e payload was launched on 21~March~2024 aboard a SpaceX cargo resupply mission and delivered to the ISS, where it was installed on the Bartolomeo external platform in August~2024. Following installation on the ISS, the payload completed standard commissioning and functional verification, after which it entered its operational phase and began executing containerised workloads for onboard AI experimentation. The compute module remains fully operational and continues to support tasking, performance evaluation, and downlink of inference metadata. All results presented in this work are obtained using both the IMAGIN-e engineering model on the ground and the flight model hosted on the ISS, which share the same configuration and resource constraints.

\subsection{Knowledge distillation for satellite deployment}\label{method1}
\subsubsection*{Teacher model} 
We adopt Prithvi-EO-2.0-300M~\cite{szwarcman2024prithvi}, a 300M-parameter Vision Transformer (ViT)~\cite{dosovitskiy2020image}-based foundation model designed for EO tasks, as our teacher model in our knowledge distillation framework~\cite{hinton2015distilling}. Its architecture (Supplementary Fig.~4) consists of $L$ transformer blocks, each comprising a multi-head self-attention (MSA) layer and an MLP block. Both blocks are preceded by layer normalisation and followed by a residual connection---that is, the output of each block is added back to its input. The MSA block projects input tokens into query, key, and value vectors, which are distributed across multiple attention heads. Each head independently performs scaled dot-product attention. The outputs from all heads are concatenated and linearly projected back to the original embedding dimension, allowing the model to capture global contextual relationships. The MLP block comprises two fully connected layers with a GeLU activation in between, and dropout is applied after each layer for regularisation. 

Before entering the encoder, the input undergoes the following preprocessing steps:
\begin{enumerate}
    \item The input data cube has dimensions ($C \times T \times H \times W$), where:
    \begin{itemize}
        \item $C$ is the number of spectral bands,
        \item $T$  is the temporal depth, representing the number of time steps (often 1 in our setting),
        \item $H \times W$ is the height and width of the input (e.g., $224 \times 224$).
    \end{itemize}
    The data cube is divided into non-overlapping patches of equal size. For a patch size of $1 \times 16 \times 16$, this results in $N$ patches, where: $N = (T / 1) \times (H / 16) \times (W / 16) = 1 \times 14 \times 14 = 196$.

    \item Each patch is flattened into a one-dimensional vector and linearly projected into a fixed-dimensional embedding space of size $D$ (e.g., 1024), producing patch embeddings of shape $(N \times D)$.

    \item A learnable class token of shape $(1 \times D)$ is prepended to the patch embeddings, resulting in a shape of $((N + 1) \times D)$.

    \item A fixed three-dimensional sine-cosine positional encoding of shape $(N + 1) \times D$ is then added to the combined patch embeddings and class token to retain spatial and temporal structure.
\end{enumerate}

The encoder outputs token embeddings of shape $(N + 1) \times D$, matching the shape of the input to the encoder. The first token corresponds to the encoded class token, and the remaining $N$ tokens represent the encoded patches. These embeddings serve as general-purpose features for downstream tasks. For segmentation, the patch tokens (excluding the class token) are reshaped into 2D feature maps and passed to a decoder head for per-pixel prediction. For classification, only the class token is used and passed to a classifier head.


\subsubsection*{Student model}
To obtain a compact variant of Prithvi, we considered several architectural parameters that influence model size: patch embedding dimension ($D$), number of encoder layers, number of attention heads, MLP hidden dimension (via the MLP ratio), and patch size. In this work, we reduced only the embedding dimension $D$, halving and quartering it from 1024 to 512 and 256, respectively. As shown in Supplementary Table~1, this results in approximately 4$\times$ and 16$\times$ reductions in the number of weights and model size, respectively, while preserving the core architectural structure of the original model.


Other parameters were kept unchanged for the following reasons:
\begin{itemize}
\item \textbf{Encoder layers:} Reducing depth alters the model's representational capacity and can destabilise the distillation process.
\item \textbf{Attention heads:} Fewer heads reduce the diversity of learned patterns, weakening the effectiveness of self-attention.
\item \textbf{MLP ratio:} Lowering this has limited size benefits but may degrade expressive power and affect performance.
\item \textbf{Patch size:} Larger patches reduce spatial granularity, hindering pixel-level precision, critical for EO applications.
\end{itemize}

While a smaller $D$ reduces representational capacity per token, this trade-off is mitigated by distillation from the larger teacher model, which guides the student to learn effective representations. This approach offers a simple and scalable strategy to reduce model size for onboard deployment without requiring extensive architectural modifications.

\subsubsection*{Pretraining via dual-MAE distillation}
Prithvi was originally pretrained on the Harmonised Landsat–Sentinel-2 (HLS)~\cite{claverie2018harmonized} dataset, which provides temporally and spectrally consistent surface reflectance products from Landsat 8/9~\cite{roy2014landsat, knight2014landsat} and Sentinel-2~\cite{drusch2012sentinel} (excluding Antarctica), with a revisit time of 2–3 days and 30~m spatial resolution. HLS includes two products:
\begin{itemize}
\item \textbf{L30:} Derived from Landsat 8/9, with 10 spectral bands spanning 0.43–12.51~$\mu$m across the visible to thermal infrared (TIR) range
\item \textbf{S30:} Derived from Sentinel-2, with 13 spectral bands spanning 0.44–2.19~$\mu$m across the visible to shortwave infrared (SWIR) range
\end{itemize}

Both L30 and S30 are processed using a common pipeline, including atmospheric correction, geometric alignment, view angle normalisation, cloud masking, and spectral harmonisation. For compatibility with the Kanyini and IMAGIN-e mission, we selected four overlapping bands---RGB and NIR---as input channels for pretraining. These bands are consistently available across HLS, Kanyini, and IMAGIN-e, and exclude coastal aerosol and SWIR bands not supported onboard. 

The compact models---Prithvi-EO-2.0-512 and Prithvi-EO-2.0-256---were pretrained on HLS using self-supervised learning via MAE~\cite{he2022masked}, consistent with the teacher model’s pretraining strategy. However, rather than reconstructing the original input, the student was supervised to match the teacher’s output. As illustrated in Supplementary Fig.~5, the teacher and student each use a separate MAE with independent encoder--decoder pipelines. The student is trained to align its reconstructions with those of the teacher at masked locations, enabling effective knowledge transfer despite reduced model capacity. 


\vspace{0.75em}
\textbf{Student workflow:}
\begin{enumerate}
    \item Input data cubes are preprocessed as in Section~\ref{method1}, with a 75\% random patch masking rate. Only unmasked patch embeddings are passed to the encoder.
    \item Masked patches are replaced by learnable tokens in the feature space.
    \item Positional encodings are added to all tokens before being passed to the decoder.
    \item The decoder reconstructs all patches (masked and unmasked).
\end{enumerate}

\textbf{Teacher workflow:}
\begin{enumerate}
    \item The same input is passed without masking (0\% mask ratio); all patch embeddings are sent to the encoder.
    \item The decoder reconstructs the full data cube, serving as the reference output.
\end{enumerate}

The mean squared error (MSE) is computed between the student's reconstructed patches at masked positions and the corresponding teacher outputs. Only the student model is updated during training; the teacher remains frozen. After pretraining, the decoder is discarded and the student encoder is retained for evaluation on downstream tasks.

\subsection{Evaluation on downstream EO tasks}\label{method2}
Following pretraining, we evaluated all Prithvi-EO-2.0 encoder configurations on five downstream tasks spanning classification and segmentation, each corresponding to a distinct EO application. In every case, the encoder weights were frozen while task-specific heads were connected to the encoder and fine-tuned on labelled datasets. For each task, we outline the dataset, model architecture, and training procedure, with a summary of head architectures, number of weights, and memory footprints provided in Supplementary Table~2.

\subsubsection*{Cloud detection}
The Sentinel-2 Cloud Mask Catalogue~\cite{francis_alistair_2020_4172871} contains cloud masks for 513 Sentinel-2 Level-1C top-of-atmosphere (TOA) reflectance data cubes ($1024 \times 1024$ pixels), collected across diverse geographical regions. The masks are binary, with $0$ denoting not-cloudy and $1$ denoting cloudy. Each data cube includes 13 spectral bands in the VNIR range and was resampled to a spatial resolution of 20~m (if not already at that resolution) using bilinear interpolation. The cubes were then divided into 2052 subcubes of size $512 \times 512$ pixels. The cloud masks served as ground-truth labels to train a segmentation head for pixel-level cloud detection. In addition, a classification head was trained for tile-level cloud detection, assigning a binary label (\emph{cloudy} or \emph{not cloudy}) based on the proportion of cloudy pixels per tile. A threshold of 70\% was applied: tiles with more than 70\% cloud coverage were labelled as \emph{cloudy}, while the rest were labelled as \emph{not cloudy}. The resulting dataset was then split into training and testing sets. A sample Sentinel-2 RGB composite and its corresponding cloud mask are shown in Supplementary Fig.~6.


For classification, a multi-layer perceptron (MLP) was employed as the classifier head. The MLP comprises two fully connected layers with ReLU activations. It operates on the class token produced by the encoder and outputs a binary label indicating whether the data cube is \emph{cloudy} (1) or \emph{not cloudy} (0). For segmentation, a UNet-style decoder~\cite{ronneberger2015u} was employed as the segmentation head. The architecture consists of four upsampling stages. In the first three stages, each transposed convolution is followed by a skip connection and a pair of convolutional layers with batch normalisation and ReLU activation. Skip connections are formed from intermediate features of four encoder layers, which are first projected into 2D feature maps using $1 \times 1$ convolutions and then upsampled with transposed convolutions to match the current spatial resolution. Dropout is applied after each of the first three stages. In the final stage, a transposed convolution followed by a $1 \times 1$ convolution produces the segmentation logits. The decoder input consists of patch tokens from multiple ViT layers (excluding the class token), reshaped into spatial feature maps before generating a mask at the original input resolution. 

Both the segmentation and classification heads were trained using supervised learning with a weighted cross-entropy loss. To penalise false positives---and thereby reduce the risk of discarding useful (non-cloudy) data during onboard filtering---a class weight ratio of 2:1 was applied to the non-cloudy and cloudy classes, respectively. Training was conducted using the Adam optimiser~\cite{kingma2014adam} for 60 epochs with a batch size of 32 for segmentation, and 300 epochs with a batch size of 128 for classification. A learning rate schedule was employed that increased linearly during the initial phase and then decayed following a cosine annealing schedule. To prevent overfitting, early stopping based on validation loss was applied. Both models were trained using horizontal and vertical flips as data augmentation.

\subsubsection*{Flood Detection}
Sen1Floods11~\cite{bonafilia2020sen1floods11} dataset contains water masks for 446 flood scenes, where each scene includes Sentinel-1 2D SAR backscatter images and corresponding Sentinel-2 Level-1C TOA reflectance data cubes (512 $\times$ 512 pixels). The masks assign a value of -1 for missing data, 0 for non-water pixels and 1 for water pixels. The dataset spans 11 flood events from 2018 to 2020, covering 14 biomes, 357 ecoregions, and 6 continents. Each Sentinel-2 data cube contains 13 spectral bands, which were resampled to a uniform spatial resolution of 10 m, if not already at that resolution. These water masks were used as ground truth labels to train a segmentation head for pixel-level water detection. A sample Sentinel-2 RGB composite and its corresponding flood mask are shown in Supplementary Fig.~7.


A UPerNet-style decoder~\cite{xiao2018unified} was employed as the segmentation head. The architecture consists of a Pyramid Pooling Module (PPM), a Feature Pyramid Network (FPN), and a final upsampling stage. The PPM operates on the top-level feature map by applying multi-scale average pooling at different grid sizes, projecting the pooled features with $1 \times 1$ convolutions, and upsampling them back to the original resolution using transposed convolutions, thereby capturing global contextual information. The FPN refines features from all four encoder layers through lateral $1 \times 1$ convolutions, followed by convolutional layers with batch normalisation and ReLU activation. Each feature is upsampled to a common spatial size and fused into a unified representation. The decoder input consists of patch tokens from multiple ViT layers (excluding the class token), which are reshaped into 2D spatial feature maps before producing a segmentation mask at the original input resolution. 

The segmentation head was trained using supervised learning with a weighted cross-entropy loss, where a class weight ratio of 1:4 was applied to the non-water and water classes, respectively. This weighting scheme was used to address class imbalance and mitigate false negatives in flood regions. Training was conducted over 300 epochs with a batch size of 32, using the Adam optimiser. The learning rate was scheduled to linearly increase during the initial phase of training, followed by gradual decay via cosine annealing. Early stopping based on validation loss was used to prevent overfitting. To enhance generalisation and increase data variability, random horizontal and vertical flips were applied as part of the data augmentation pipeline. Given the moderate class imbalance in the dataset (10\% water, 77\% non-water, 13\% missing), we report detailed performance on the water class to assess model capability in detecting flooded regions.

\subsubsection*{Landslide Detection}
Landslide4Sense~\cite{ghorbanzadeh2022outcome,ghorbanzadeh2022landslide4sense} dataset contains landslide masks for 4,884 Sentinel-2 data cubes ($128 \times 128$ pixels), collected from diverse mountainous regions across the globe between 2015 and 2021. The masks assign a value of 0 to non-landslide pixels and 1 to landslide pixels. Each data cube includes 12 spectral bands in the VNIR range, resampled to a spatial resolution of 10 m (if not already at that resolution). Although the dataset does not explicitly specify the processing level---whether Level-1C (TOA) or Level-2A (surface reflectance)---it is widely used as a benchmark for landslide detection. The dataset also includes slope information and digital elevation model (DEM) data derived from the Advanced Land Observing Satellite Phased Array type L-band Synthetic Aperture Radar (ALOS PALSAR)~\cite{shimada2009palsar,takaku2014generation}. These landslide masks were used as ground truth labels to train a segmentation head for pixel-level landslide detection. A sample Sentinel-2 RGB composite and its corresponding landslide mask is shown in Supplementary Fig.~8.


The UNet-style decoder, previously used for cloud detection, was employed as the segmentation head. It was trained using supervised learning with a weighted cross-entropy loss, where a class weight ratio of 1:4 was applied to the non-landslide to landslide classes, respectively. This weighting scheme was used to address the extreme class imbalance and mitigate false negatives in landslide regions. Training was conducted over 150 epochs with a batch size of 32, using the Adam optimiser. A cosine annealing learning rate schedule with warm restarts was applied, together with early stopping based on validation loss to prevent overfitting. Data augmentation consisted of random horizontal and vertical flips. Given the extreme imbalance in the dataset (approximately 98\% non-landslide and 2\% landslide pixels), evaluation focused on reporting detailed performance metrics for the landslide class to assess the model's ability to detect rare events.

\subsubsection*{AGB estimation}
BioMassters~\cite{nascetti2023biomassters} dataset contains AGB estimation masks for 11,463 scenes, where each scene includes Sentinel-1 2D SAR backscatter images and corresponding Sentinel-2 Level-2A BOA reflectance data cubes (256 $\times$ 256 pixels). The AGB masks were generated from airborne LiDAR campaigns conducted by the Finnish Forest Centre and the National Land Survey of Finland, which combined dense LiDAR point clouds ($\geq$ 5 points/m$^2$), high-resolution aerial imagery (0.4 m GSD), and in-situ field plots, using calibrated allometric models to derive AGB values per 10 m $\times$ 10 m pixel. The resulting masks report AGB in tonnes per hectare (t/ha), with zero values indicating missing data or no biomass. Each Sentinel-2 data cube includes 10 spectral bands (B1, B9, B10 not used) in the VNIR range, resampled to a spatial resolution of 10 m (if not already at that resolution). A sample Sentinel-2 RGB composite and its corresponding AGB ground truth map is shown in Supplementary Fig.~9.


The UPerNet-style decoder, previously used for flood detection, was employed as the segmentation head. It was trained using supervised learning with a per-image root mean squared error (RMSE) loss, where zero-valued pixels were masked to avoid bias from missing data. Training was conducted over 200 epochs with a batch size of 20, using the Adam optimiser. A cosine annealing learning rate schedule with warm restarts was applied, together with model selection based on validation RMSE to prevent overfitting. Data augmentation consisted of random cropping to $224 \times 224$ patches and random horizontal and vertical flips, with centre cropping applied during validation and testing. Model evaluation focused on reporting RMSE to quantify the accuracy of AGB estimation.




\subsection{Domain adaptation to target domains}\label{method3}
To ensure generalisation to real mission data, we performed domain adaptation using imagery captured from the Kanyini mission. Specifically, we selected the most compact and highest-performing variant of Prithvi as our GeoFM. Task-specific heads, initially fine-tuned on labelled benchmark datasets in Section~\ref{method2}, were subsequently adapted to the Kanyini domain. This two-stage process was introduced under the assumption that a domain gap exists between the benchmark datasets (source domain) and imagery from Kanyini (target domain). By contrast, no domain adaptation was possible for the IMAGIN-e mission because the imager suffered a power fault and produced no operational data products.

We next describe the Kanyini data products and the preprocessing pipeline used to generate inputs for model training. We then outline the ground-truth generation procedure for flood and cloud detection, followed by the adaptation strategy applied to Kanyini data, including class balancing and sampling schemes.

\subsubsection*{Kanyini data products}\label{kanyini-data}
Kanyini data products (Supplementary Fig.~10) consist of Level-1 TOA reflectance data cubes with 50 spectral bands, covering the VNIR range (0.45--0.95~$\mu$m), at a spatial resolution of 75~m. These products were generated by processing Level-0 frames---individual, unstructured images---using a proprietary image processing tool provided by the camera manufacturer. The processing pipeline includes radiometric and geometric corrections, as well as cube assembly (via stacking and spatial alignment) to produce Level-1 TOA hyperspectral data cubes.

A total of 40 data products were selected. Four bands, centred at 0.49~$\mu$m, 0.60~$\mu$m, 0.66~$\mu$m, and 0.87~$\mu$m, were used to represent the RGB and NIR channels. TOA reflectance values less than zero were clipped to zero. Missing values (encoded as 32,767) were also set to zero, and all values were normalised to the range $[0, 1]$. Each data cube was divided into 88 non-overlapping sub-cubes of size $224 \times 224$ pixels, yielding a total of 3,520 tiles. To better reflect operational conditions, the dataset was split into training and testing sets based on data products rather than individual tiles. The specific split varied depending on the downstream task. 

\subsubsection*{Ground truth generation}\label{annotation}
We employed automated labelling techniques to generate proxy ground truth masks for model training and evaluation. Surface water was extracted using the \textit{Normalized Difference Water Index} (NDWI)~\cite{mcfeeters1996ndwi}, while a cloud index was computed using the \textit{Haze Optimized Transform} (HOT)~\cite{zhang2002image}, following the adaptation described in \cite{zilberstein2024demonstrating}. The NDWI is computed as:

\begin{equation}
\text{NDWI} = \frac{R_{\text{green}} - R_{\text{NIR}}}{R_{\text{green}} + R_{\text{NIR}}}
\end{equation}

where \( R_{\text{green}} \) and \( R_{\text{NIR}} \) are the reflectance values in the green and near-infrared bands, respectively. The resulting NDWI image is thresholded using Otsu’s method~\cite{otsu1979threshold} to produce a binary surface water mask (Supplementary Fig.~10a) for each scene. 

The HOT method leverages the spectral relationship between blue and red band reflectance values under clear-sky conditions. The red band is typically less affected by atmospheric scattering, while the blue band exhibits greater sensitivity to haze and cloud interference. In clear-sky pixels, these two bands follow a well-defined linear correlation, referred to as the \textit{clear-sky line}. Pixels that deviate significantly from this line are likely influenced by atmospheric contamination. To estimate the clear-sky line, we select the 0.15\% of pixels with the lowest blue reflectance. These are divided into 20 bins, and from each bin, the 20 pixels with the highest red reflectance are retained. A linear regression is then applied to the resulting 400 points, yielding a slope \( m \) and intercept \( b \) that define the clear-sky line. The HOT value for each pixel is computed as:

\begin{equation}
\text{HOT} = |m \cdot R_{\text{blue}} - R_{\text{red}}| + \frac{b}{\sqrt{1 + m^2}}
\end{equation}

where \( R_{\text{blue}} \) and \( R_{\text{red}} \) are the reflectance values in the blue and red bands, respectively. As with NDWI, the resulting cloud index is thresholded using Otsu’s method to generate a binary cloud mask (Supplementary Fig.~10b). Both the surface water and cloud labels are manually verified and, where necessary, corrected to ensure high-quality annotations. 


At this stage, only water and cloud masks were generated. Other downstream tasks, such as landslide detection and AGB estimation, were not attempted due to the absence of suitable ground truth data. Landslide masks could not be produced, as no visible landslide events were present in the dataset. AGB estimation typically requires spatially co-located reference maps derived from field plots, airborne LiDAR, or calibrated satellite-based biophysical models, none of which were available for the spatial and temporal extent of the Kanyini scenes. 







\subsubsection*{Adaptation to the Kanyini domain}\label{adaptation-kanyini}
We selected task-specific subsets from a pool of 40 available Kanyini data products. A total of 33 products were used to train task heads for cloud detection (both tile-level classification and pixel-level segmentation), and 30 products were selected for flood detection. Data products were chosen independently for each task to ensure sufficient representation of the target phenomena---cloud-covered scenes for cloud detection and inundated regions for flood detection---and to promote generalisation across varying environmental conditions. 

The cloud detection dataset exhibited a pronounced class imbalance, with non-cloudy tiles significantly outnumbering cloudy ones. To mitigate this, we applied different sampling strategies for tile-level and pixel-level tasks. For tile-level classification, we employed a weighted random sampling scheme to upsample cloudy tiles and achieve a balanced 1:1 ratio between classes during each training epoch. Class weights were derived from the ratio of non-cloudy to cloudy tiles in the training set. For pixel-level segmentation, we selectively downsampled clear-sky tiles based on their cloud coverage. Each tile was assigned a cloud ratio, defined as the proportion of cloud pixels in the corresponding binary mask. Tiles with near-zero cloud coverage were randomly subsampled to match the number of high-cloud tiles (those with a cloud ratio $\geq$ 0.70), while all other tiles were retained. This approach ensured that cloud pixels were adequately represented during training while reducing the dominance of clear-sky scenes. The same downsampling strategy was also applied to the flood detection dataset, which similarly exhibited a significant imbalance with far more non-water scenes, to ensure adequate representation of scenes containing water. 

To build on these task-specific adaptations and sampling strategies, we systematically evaluated the data efficiency (i.e., performance under limited supervision) and reusability (i.e., deployment without target-domain updates) of our GeoFM. We conducted experiments using varying fractions of labelled Kanyini training data: 100\%, 50\%, and 25\%. We compared three configurations: (1) a randomly initialised ViT encoder with the same architecture as our GeoFM together with a randomly initialised task head (\textit{Randomly init. ViT encoder and head}); (2) our GeoFM with a randomly initialised task head (\textit{GeoFM with randomly init. head}); and (3) our GeoFM with a task head pretrained on source-domain data (\textit{GeoFM with pretrained head}). In configurations (2) and (3), the GeoFM was kept frozen, while only the task heads were retrained. These experiments were designed to address three key questions:
\begin{enumerate}
\item Does the foundation model reduce the amount of labelled data required for effective downstream performance?
\item Can the foundation model be deployed without further adaptation?
\item Is task head pretraining necessary for effective adaptation?
\end{enumerate}

\subsection{Validation on flight-representative hardware}\label{method4}
Having performed domain adaptation, we validated our GeoFM together with its task heads on flight-representative hardware to ensure deployment readiness under realistic mission constraints. These ground-based environments replicate the computational resources and system constraints of the HyperScout-2 payload aboard Kanyini and the compute module of the IMAGIN-e payload aboard the ISS. In this context, validation refers to confirming reliable model execution, seamless integration with the onboard hardware, consistent inference performance, and efficient use of computational and power resources across flight-representative environments.

\subsubsection*{HyperScout-2 (Kanyini)}
HyperScout-2 is a compact hyperspectral and thermal imaging instrument with onboard processing capability (Supplementary Fig.~11). It integrates two 2D push-broom sensors:

\begin{itemize}
    \item \textbf{VNIR sensor (FEE-VNIR).} A CMOS-based detector capturing 50 spectral bands spanning 0.45–0.95~$\mu$m at 75~m spatial resolution. 
    
    \item \textbf{TIR sensor (FEE-TIR).} A microbolometer capturing 3 spectral bands spanning 8–14~$\mu$m at 330~m spatial resolution.
\end{itemize}

In addition, the payload comprises several processing and control subsystems: 

\begin{itemize}    
    \item \textbf{Back-end electronics (BEE).} An FPGA-based board responsible for distributing power, clock signals, telemetry and commands, as well as interfacing with data and control. It creates raw data, stores it in the mass memory unit (MMU), and hosts the CPU-based onboard data handler (OBDH). The OBDH is equipped with an Intel Atom E3825 dual-core 1.33~GHz processor and 2~GB RAM, while each MMU provides 64~GB of storage for image data. Primary tasks include data acquisition, pre/post-processing, and transfer.
 
    \item \textbf{Eyes of Things (EOT).} A low-power VPU-based board using the Intel Myriad-2 VPU to execute and accelerate onboard ML inference. It interfaces with the OBDH via USB. 
    
    \item \textbf{Instrument control unit (ICU).} A supervisory module that manages and monitors the status of all subsystems. It remains continuously powered when the payload is active, while other subsystems are powered selectively according to the operating mode. For example, in VPU processing mode, the ICU powers on the OBDH, EOT, and one MMU to enable AI inference.   
    
    \item \textbf{Spacecraft bus (S/C BUS).} Provides physical integration, power supply, thermal control, and communications for all subsystems. 
\end{itemize}


Validation of our GeoFM and its task heads was performed by our team on (i) a hardware emulator (Supplementary Fig.~12a), which provided a controlled, reconfigurable environment for debugging and iterative development, and (ii) the HyperScout-2 engineering model (Supplementary Fig.~12b), which is electrically and functionally equivalent to the flight unit but operated under laboratory conditions. The validation process began by converting the models (GeoFM with task heads) into the OpenVINO IR format. PyTorch models were first exported to ONNX, then converted to IR with quantisation from FP32 to FP16, as required by the Myriad-2. Each IR model, comprising an XML architecture file and a BIN weight file, was uploaded to the hardware emulator. The emulator replicated the onboard compute environment of HyperScout-2, consisting of a Ubotica CogniSAT-XE1 module with a Myriad-2 and an Advantech ARK-1123C embedded system, whose CPU and memory specifications exactly match those of the OBDH. The same procedures were subsequently executed on the HyperScout-2 engineering model to confirm deployment readiness under flight-representative conditions, using an identical software stack---operating system, drivers, and inference libraries---across both environments to ensure that performance differences could be attributed solely to hardware rather than software variation. 


Task-level performance was evaluated on dedicated test sets for cloud and flood detection in terms of accuracy, IoU, F1 score, false positives, and related metrics, alongside efficiency and resource utilisation statistics measured on both environments. These included inference latency per tile, end-to-end runtime, peak memory usage, and power consumption (peak, average, and total energy), all reported on a per-cube basis. Latency, runtime, and memory were obtained through runtime profiling, while power consumption was measured directly at the power supply via external current–voltage logging.


\subsubsection*{IMAGIN-e (ISS)}
IMAGIN-e comprises a hyperspectral imaging instrument (HyperScout-M), an RGB camera, and a compute module for onboard processing. As the imaging subsystem was not operational during this study, validation focused on the compute module, which is equipped with 16 ARM Cortex-A72 cores, 16~GB of RAM, and 10~GB of storage, operating without dedicated hardware acceleration. The module runs a Linux-based environment and supports containerised application deployment via the Microsoft Azure Orbital Space SDK. Applications are deployed and managed within a lightweight K3s Kubernetes cluster, which abstracts low-level spacecraft telemetry and sensor interfaces, allowing applications to be deployed and managed in a standardised way for in-orbit execution. 

Validation of our GeoFM and its task heads was also carried out on (i) a RPi~4, by our team, which provided a controlled, reconfigurable environment for debugging and iterative development, and (ii) the IMAGIN-e engineering model (Supplementary Fig.~13), by Thales Alenia Space, which is electrically and functionally equivalent to the flight unit but operated under laboratory conditions. The validation process began by converting the models into the ONNX format and packaging them, together with curated test sets and the software stack required to run inference, into a Docker image. This image was first executed on the RPi~4, which shares the same ARM CPU architecture as the IMAGIN-e compute module but with fewer cores and less memory. It was then validated on the engineering model to confirm compatibility with the target compute environment and to establish baseline performance prior to on-orbit demonstration. Task-level performance was evaluated on both environments; however, efficiency and resource utilisation statistics were not assessed on the RPi~4, as it is not an exact hardware representation of the IMAGIN-e compute module.


\subsection{On-orbit demonstration}\label{method5}
Following successful on-ground validation, the GeoFM and its task heads were deployed on the IMAGIN-e payload aboard the ISS to demonstrate on-orbit inference. The deployment sequence began with validation on a curated test set representative of typical mission scenes to verify end-to-end execution. Although live image acquisition could not be performed during the testing window due to a power fault in the imager, the uploaded dataset enabled benchmarking under genuine orbital conditions. Inference was executed onboard the compute module under ground command, with intermediate outputs and runtime logs downlinked for comparison with on-ground baselines.

The on-orbit execution verified that the uploaded models and runtime environment functioned correctly in space. Model integrity was confirmed by ensuring that the architecture, parameters, and outputs remained identical to the validated ground versions. Numerical consistency checks confirmed that the inference results produced on orbit were virtually identical to those from ground testing, differing only by negligible floating-point tolerances. These outcomes collectively confirmed the stability of the onboard compute module and established overall system readiness for future missions carrying fully operational imagers.

    

\backmatter






\bmhead{Supplementary information}
Supplementary Table~1-2 and Fig.~1-13 accompany this paper.

\bmhead{Code availability}
The code used for model training, evaluation, and deployment will be released on a public repository following publication.

\bmhead{Acknowledgements}
The authors acknowledge support from SmartSat CRC for providing datasets used for model adaptation and for facilitating validation on the hardware emulator and engineering model of the HyperScout-2 payload. The authors also thank Thales Alenia Space for enabling and providing evidences of the on-orbit demonstration on IMAGIN-e.









\bibliography{sn-bibliography}

\newpage
\includepdf[pages=-]{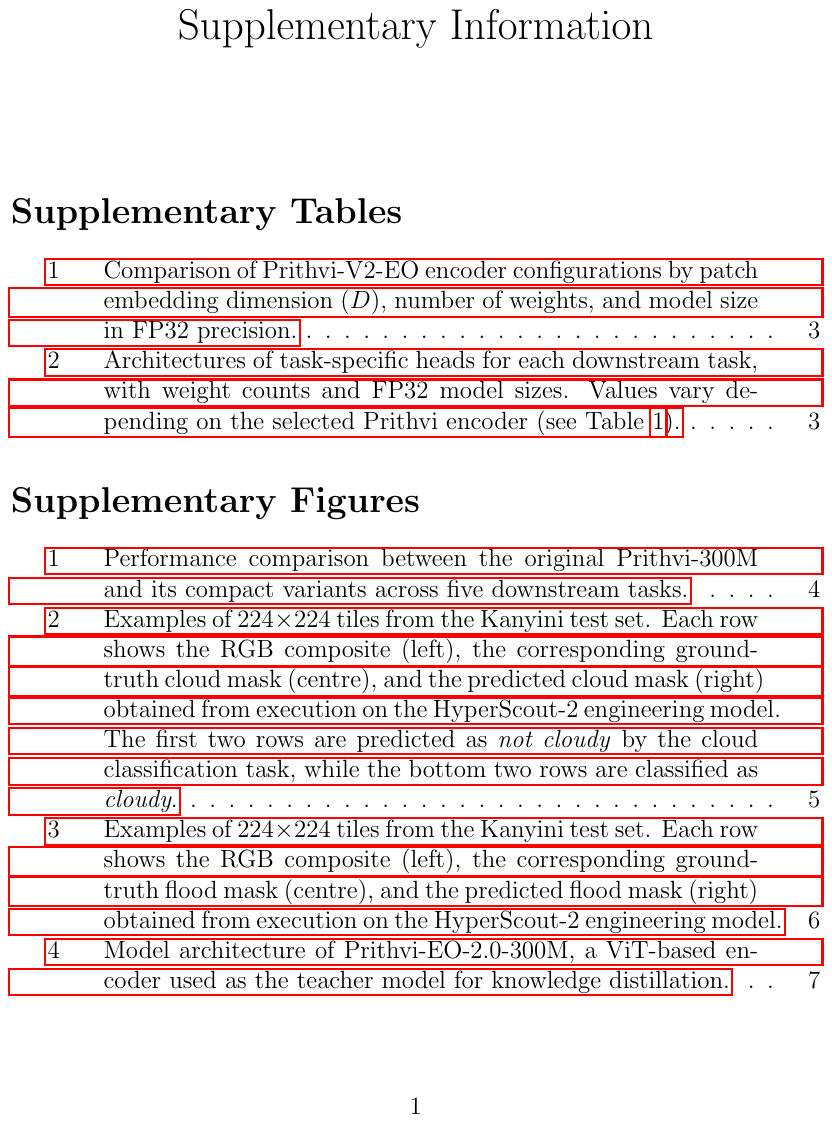}

\end{document}


\maketitle


\renewcommand{\listtablename}{Supplementary Tables}
\listoftables

\renewcommand{\listfigurename}{Supplementary Figures}
\listoffigures
\newpage



\begin{figure}[h]
  \centering
  \begin{subfigure}{0.6\textwidth}
    \includegraphics[width=\linewidth]{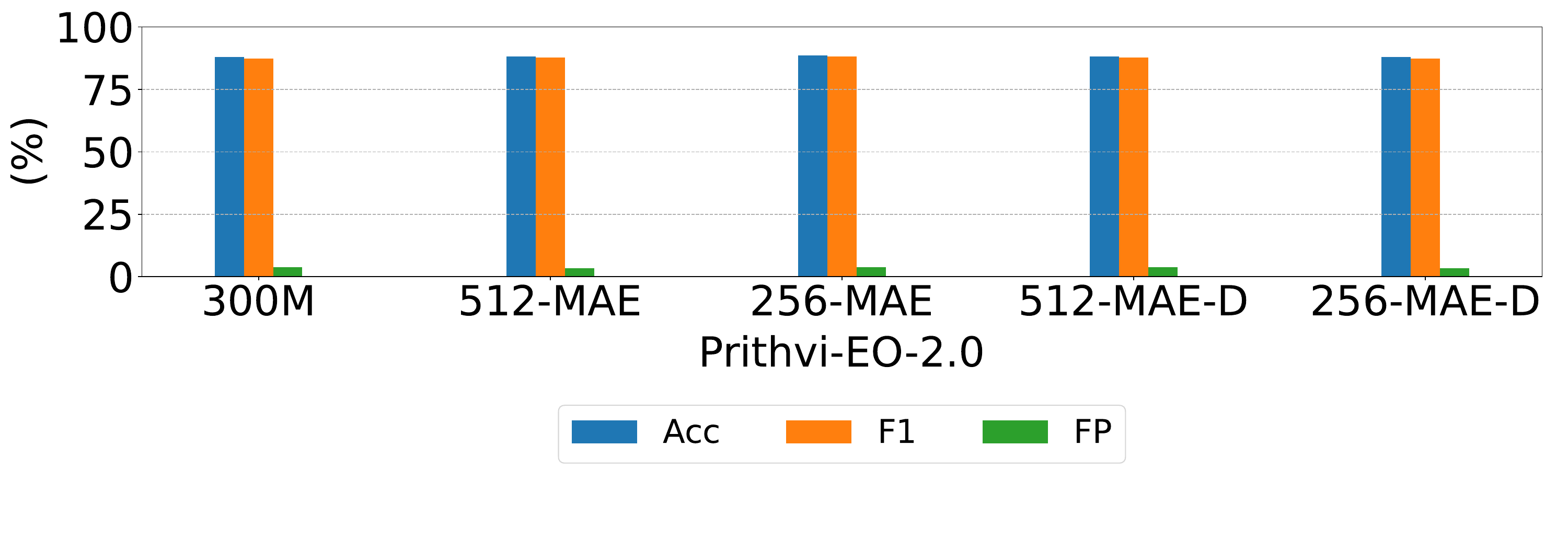}
    \vspace{-2.5em}
    \caption{Cloud detection — tile-level classification}
  \end{subfigure}

  \begin{subfigure}{0.6\textwidth}
    \includegraphics[width=\linewidth]{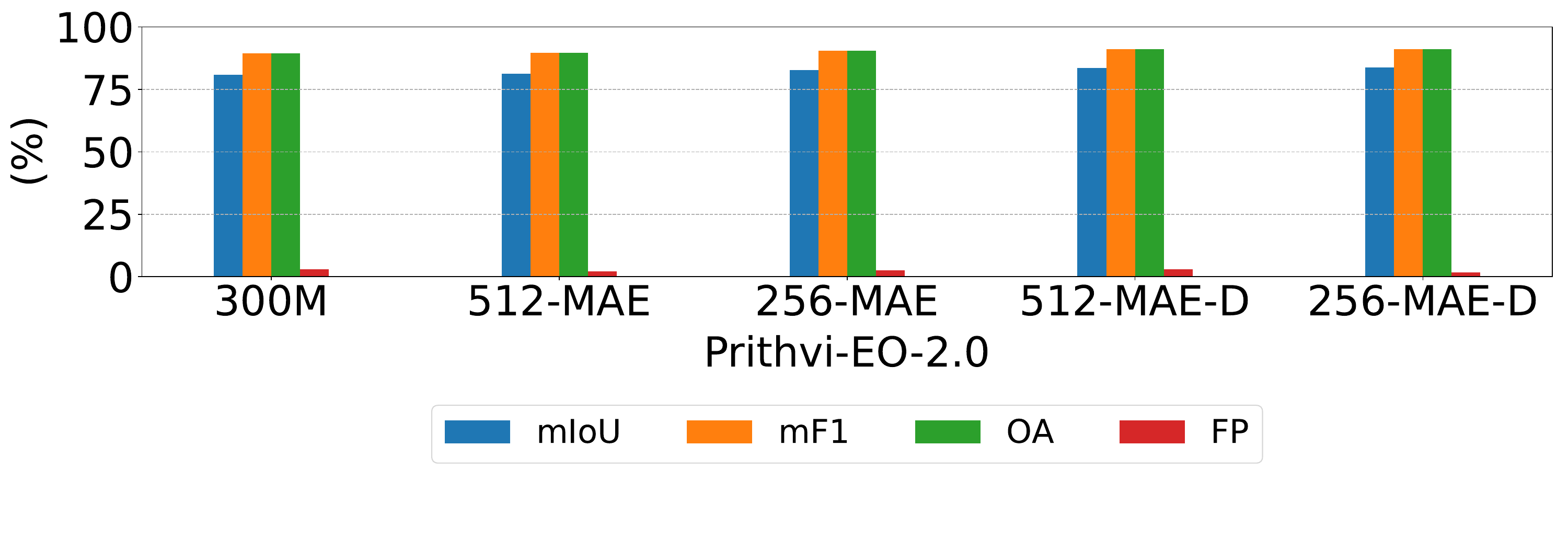}
    \vspace{-2.5em}
    \caption{Cloud detection — segmentation}
  \end{subfigure}

  \begin{subfigure}{0.6\textwidth}
    \includegraphics[width=\linewidth]{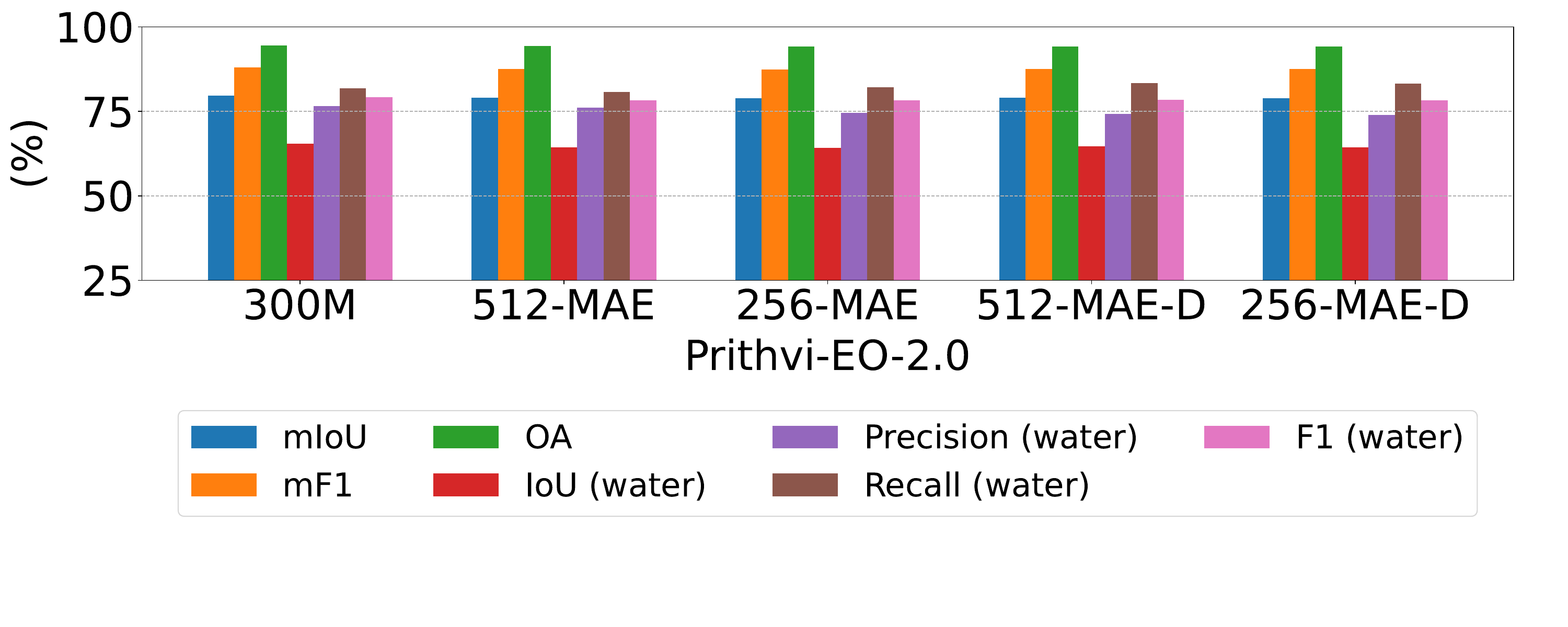}
    \vspace{-2.5em}
    \caption{Flood detection — segmentation}
  \end{subfigure}

  \begin{subfigure}{0.6\textwidth}
    \includegraphics[width=\linewidth]{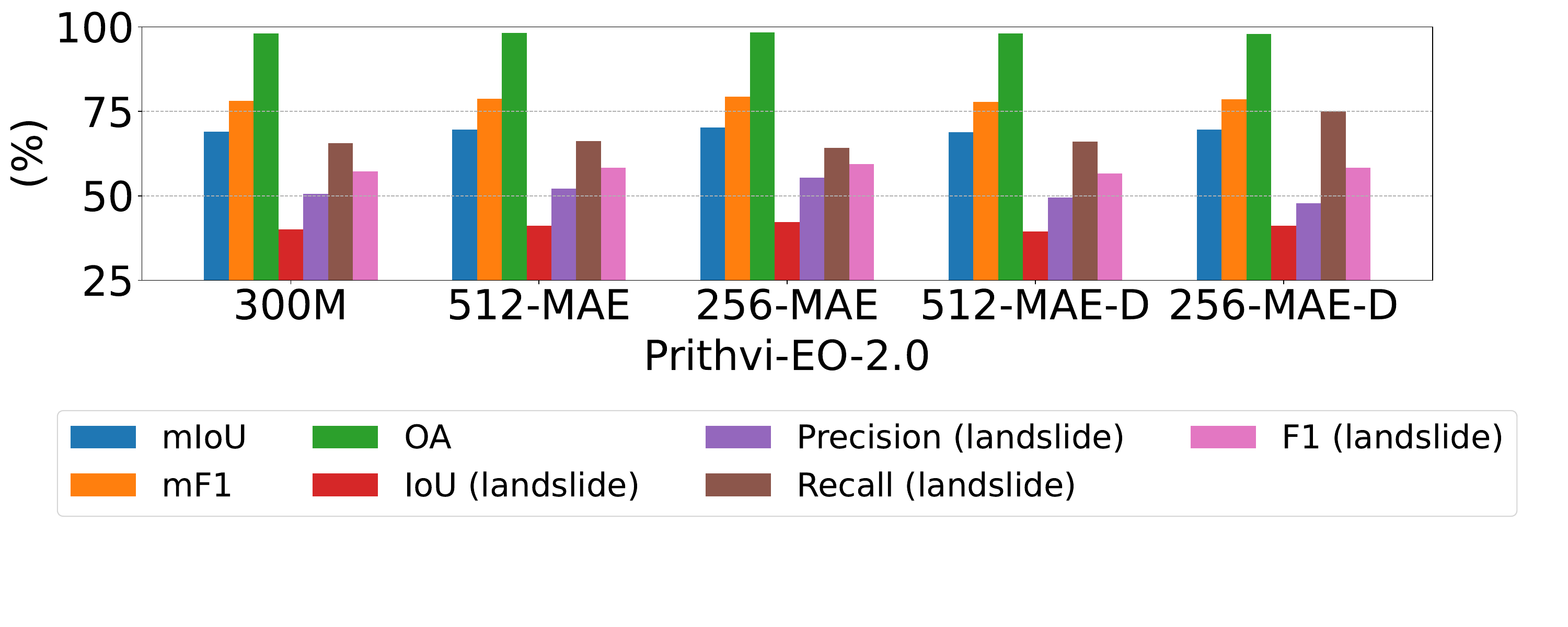}
    \vspace{-2.5em}
    \caption{Landslide detection — segmentation}
  \end{subfigure}

  \begin{subfigure}{0.6\textwidth}
    \includegraphics[width=\linewidth]{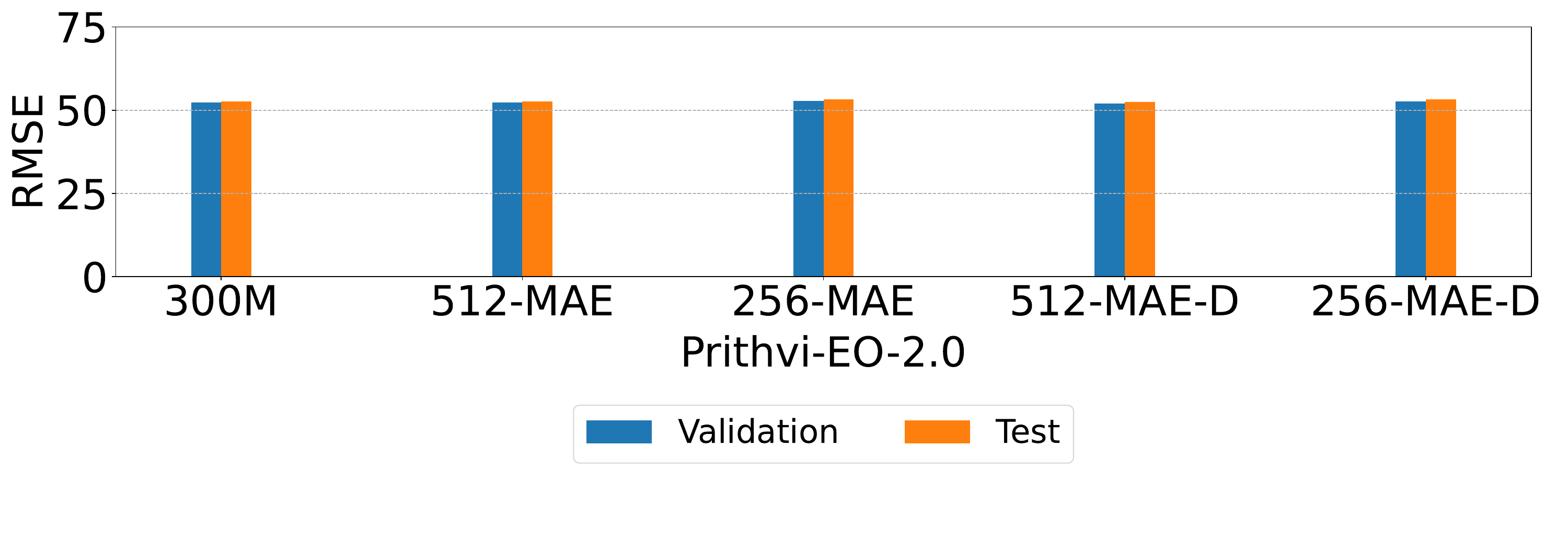}
    \vspace{-2.5em}
    \caption{AGB estimation — regression.}
  \end{subfigure}

  \caption{Performance comparison between the original Prithvi-300M and its compact variants across five downstream tasks.}
  \label{fig:supp-fig1}
\end{figure}

\begin{figure}[!htbp]
  \centering

  \begin{subfigure}{\linewidth}
    \centering
    \includegraphics[height=4.0cm]{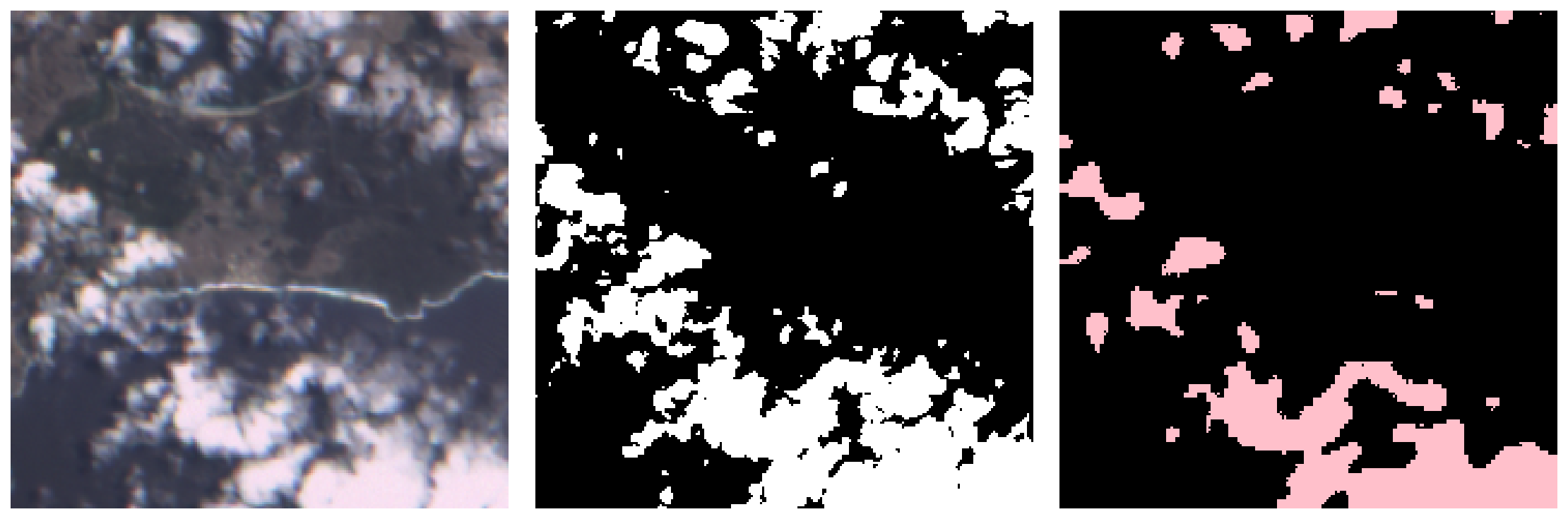}
  \end{subfigure}

  \begin{subfigure}{\linewidth}
    \centering
    \includegraphics[height=4.0cm]{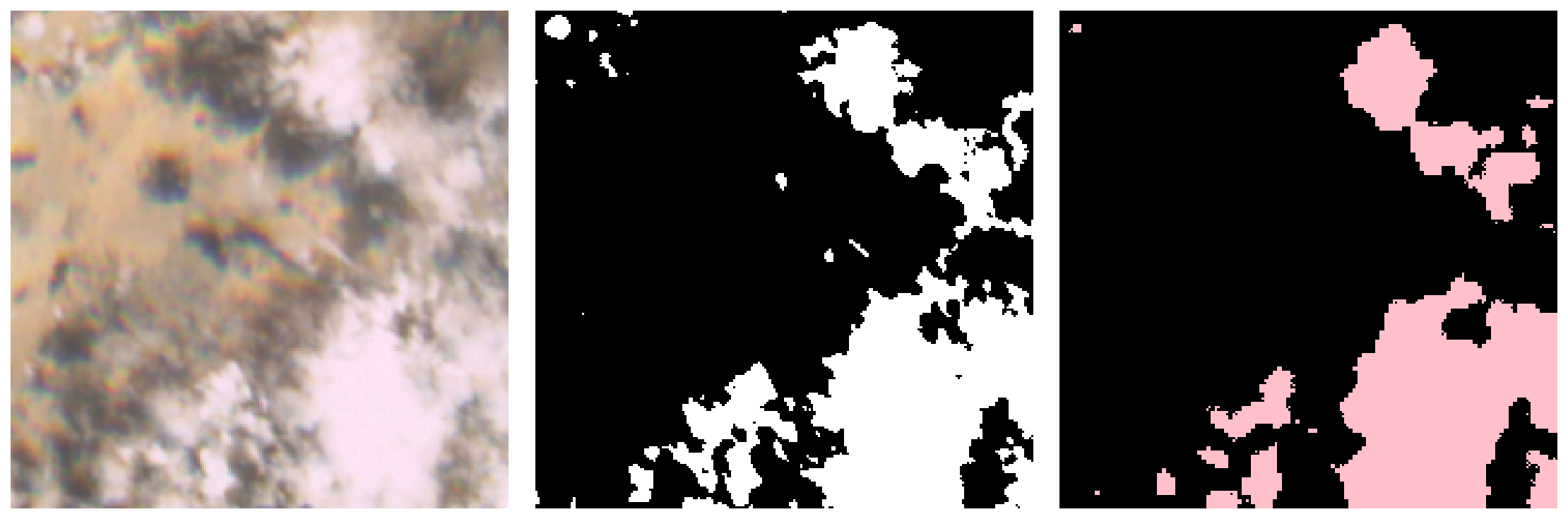}
  \end{subfigure}

  \begin{subfigure}{\linewidth}
    \centering
    \includegraphics[height=4.0cm]{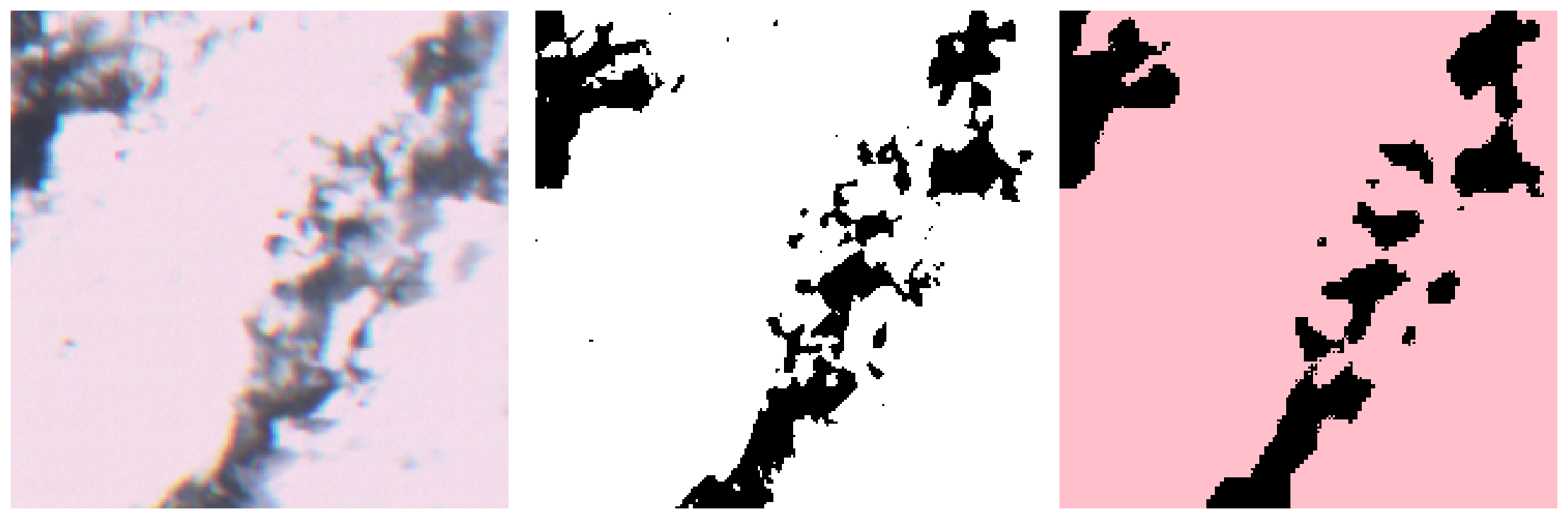}
  \end{subfigure}

  \begin{subfigure}{\linewidth}
    \centering
    \includegraphics[height=4.0cm]{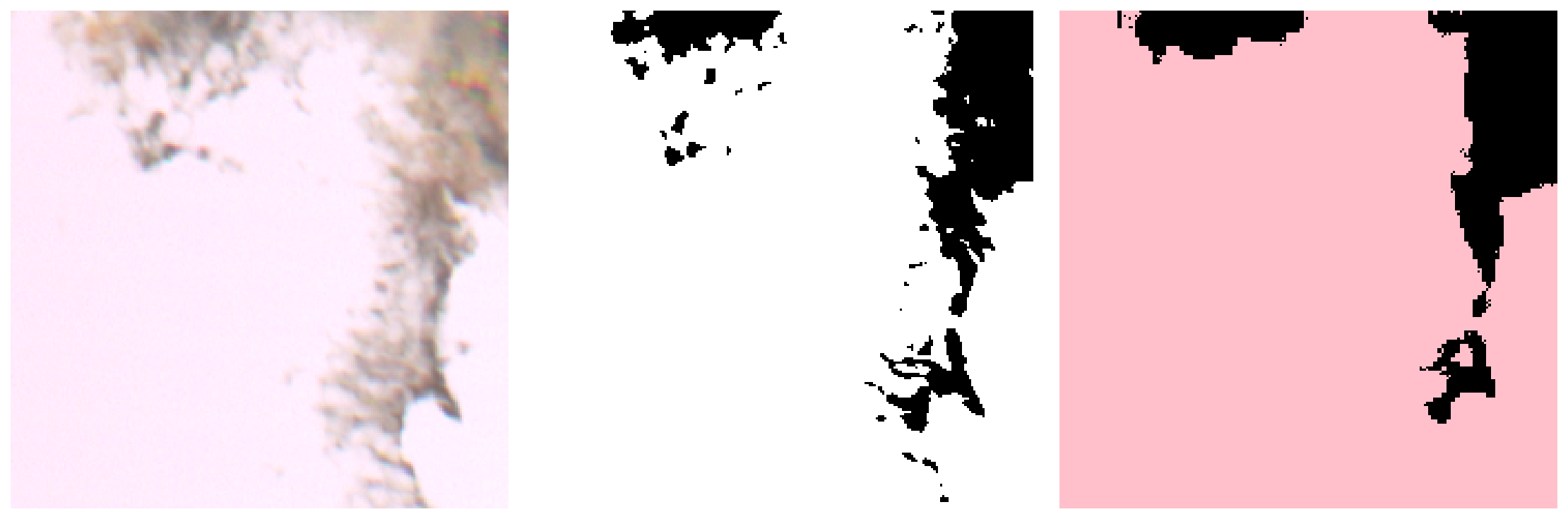}
  \end{subfigure}

  \caption{Examples of $224 \times 224$ tiles from the Kanyini test set. Each row shows the RGB composite (left), the corresponding ground-truth cloud mask (centre), and the predicted cloud mask (right) obtained from execution on the HyperScout-2 engineering model. The first two rows are predicted as \emph{not cloudy} by the cloud classification task, while the bottom two rows are classified as \emph{cloudy}.}

  \label{fig:qualitative-results-cloud}
\end{figure}

\begin{figure}[!htbp]
  \centering

  \begin{subfigure}{\linewidth}
    \centering
    \includegraphics[height=4.0cm]{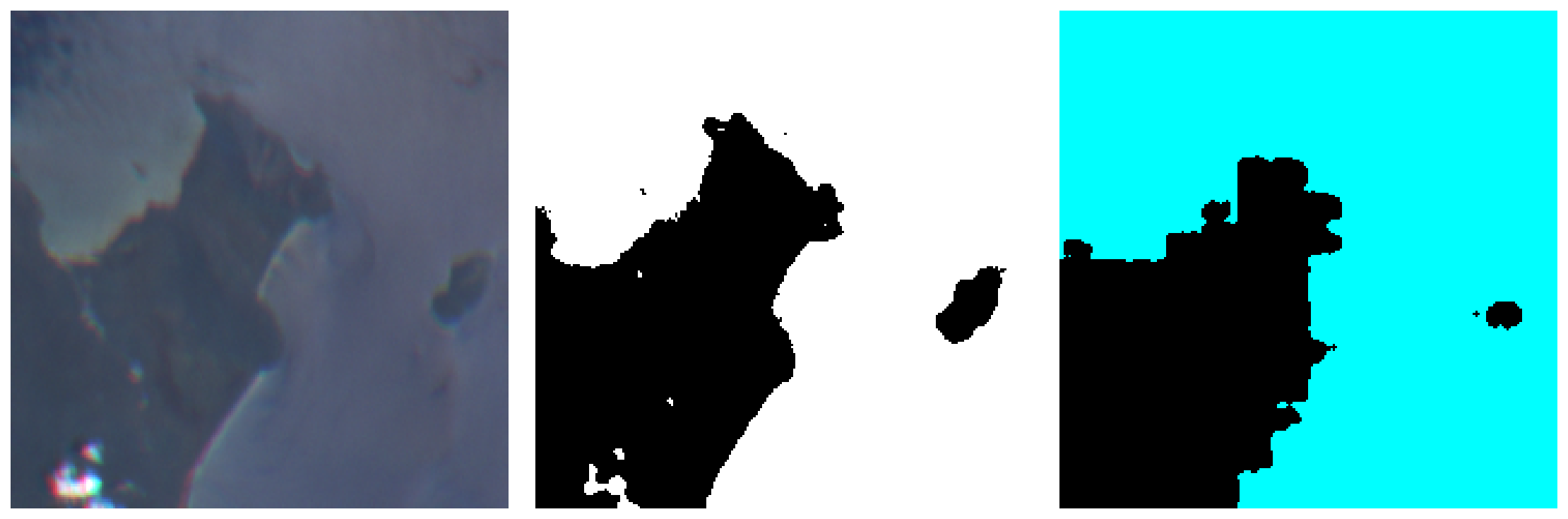}
  \end{subfigure}

  \begin{subfigure}{\linewidth}
    \centering
    \includegraphics[height=4.0cm]{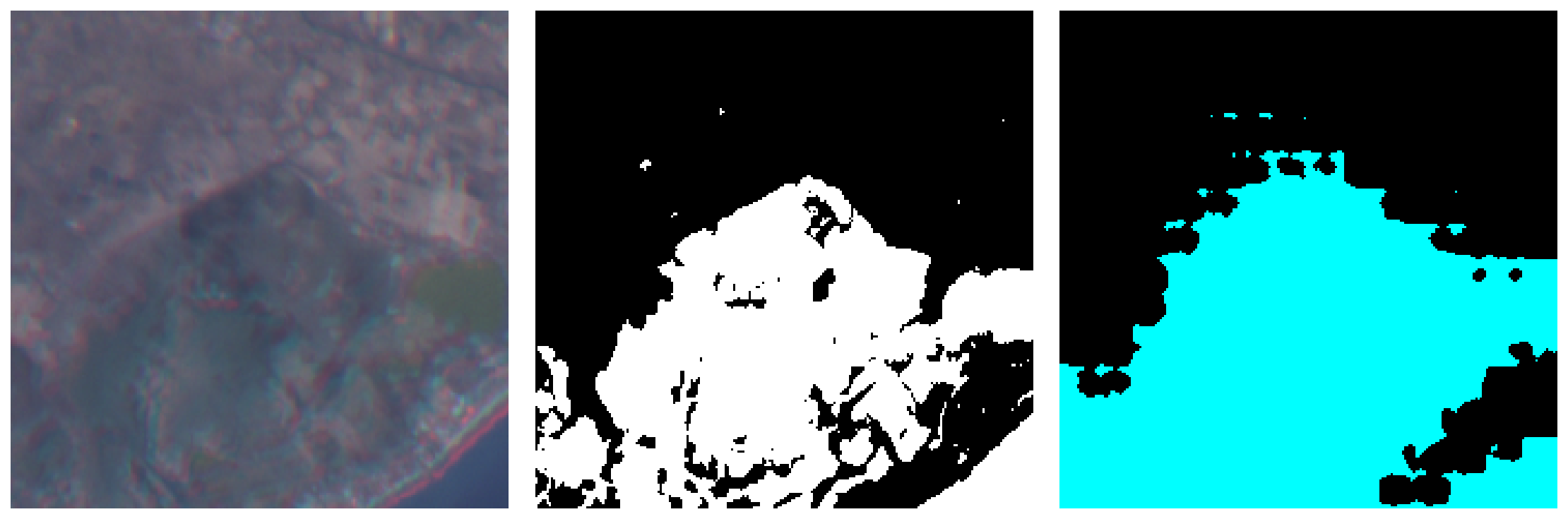}
  \end{subfigure}

  \begin{subfigure}{\linewidth}
    \centering
    \includegraphics[height=4.0cm]{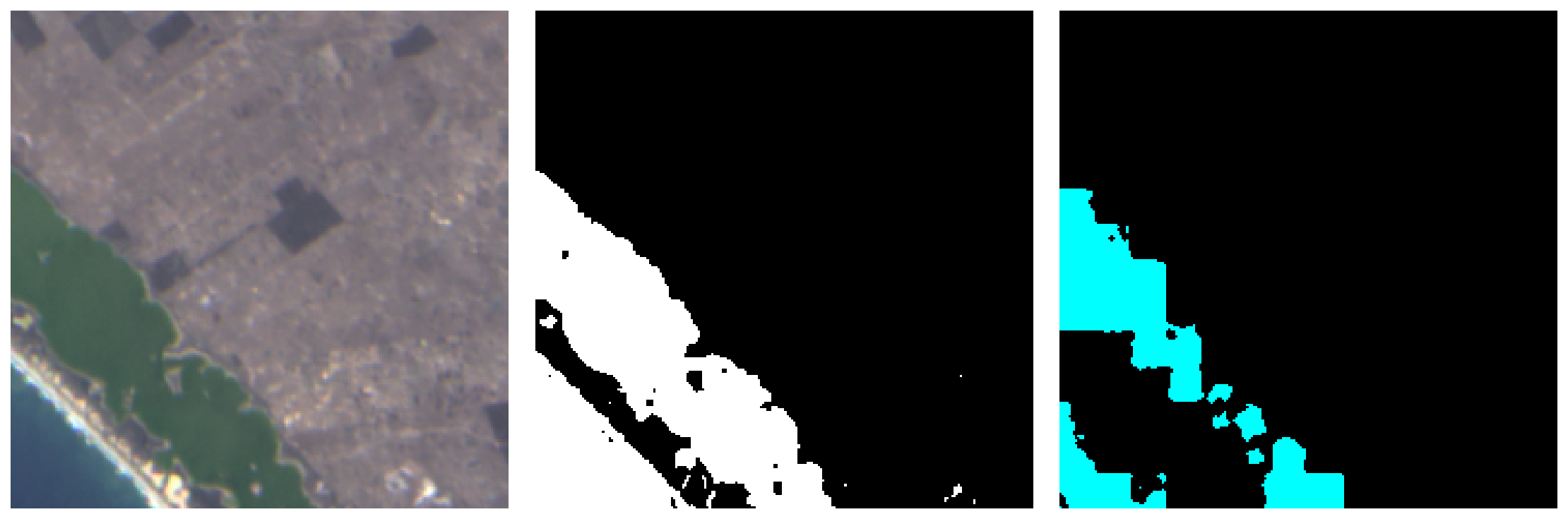}
  \end{subfigure}

  \begin{subfigure}{\linewidth}
    \centering
    \includegraphics[height=4.0cm]{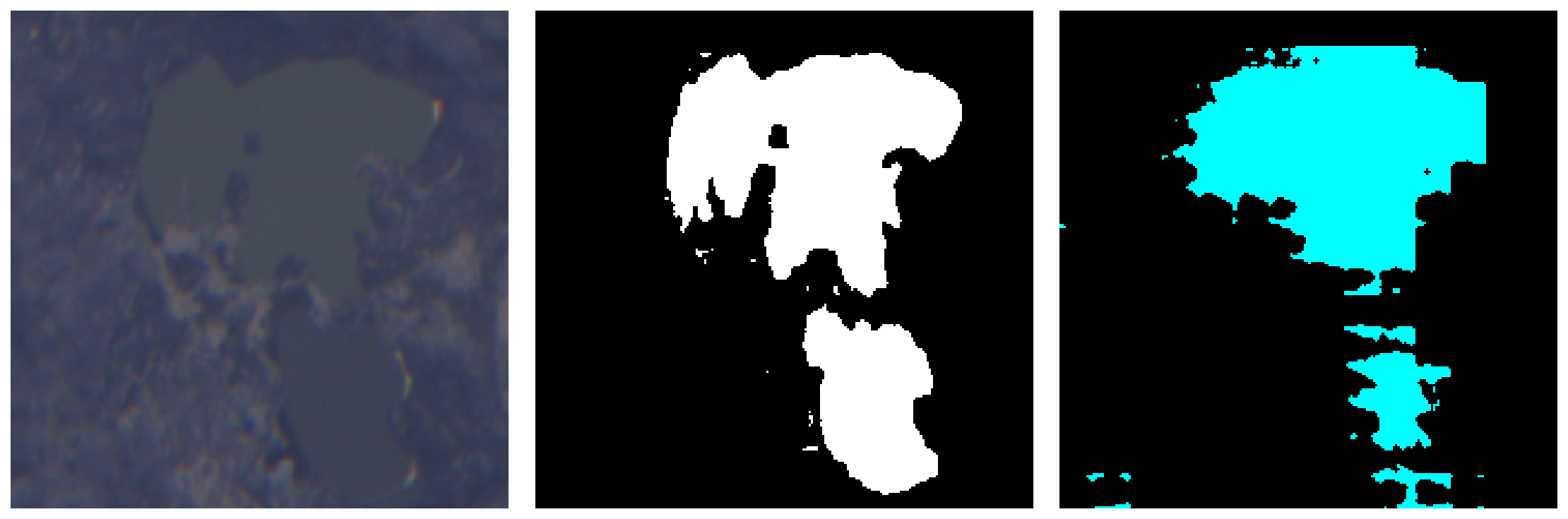}
  \end{subfigure}

  \caption{Examples of $224 \times 224$ tiles from the Kanyini test set. Each row shows the RGB composite (left), the corresponding ground-truth flood mask (centre), and the predicted flood mask (right) obtained from execution on the HyperScout-2 engineering model.}

  \label{fig:qualitative-results-water}
\end{figure}

\begin{figure}[h]
\centering
\includegraphics[width=\linewidth]{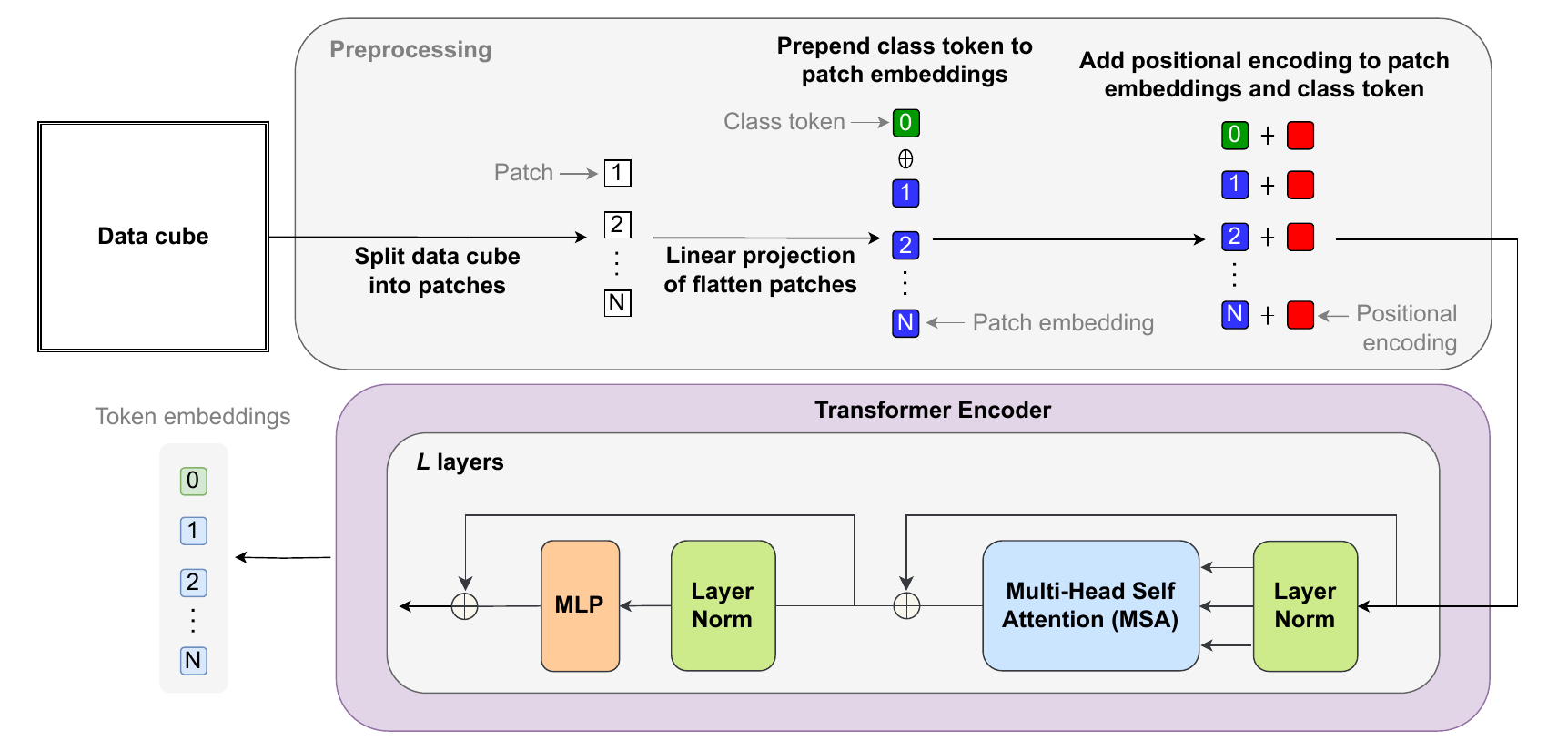}
\caption{Model architecture of Prithvi-EO-2.0-300M, a ViT-based encoder used as the teacher model for knowledge distillation.}
\label{fig:prithvi-v2-architecture}
\end{figure}

\begin{figure}[h]
\centering
\includegraphics[width=\linewidth]{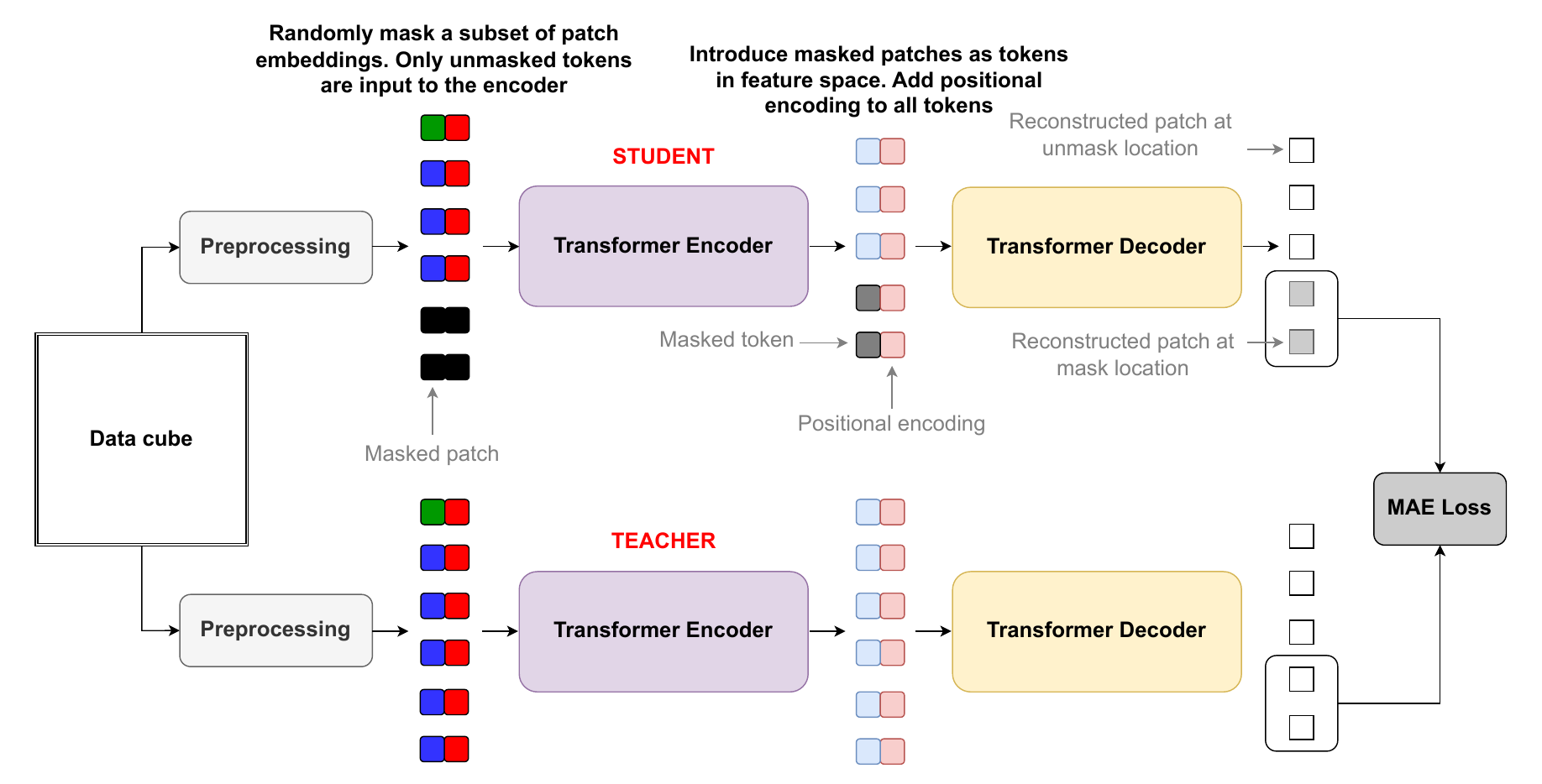}
\caption{Dual-MAE distillation setup for pretraining compact variants of Prithvi-EO-2.0-300M. The teacher and student each employ independent MAE encoder–decoder pipelines; the student is trained to match the teacher’s reconstructions at masked patch locations, enabling knowledge transfer despite reduced embedding dimension.}
\label{fig:mae-distillation}
\end{figure}

\begin{figure}[h]
\centering
\includegraphics[width=0.90\linewidth]{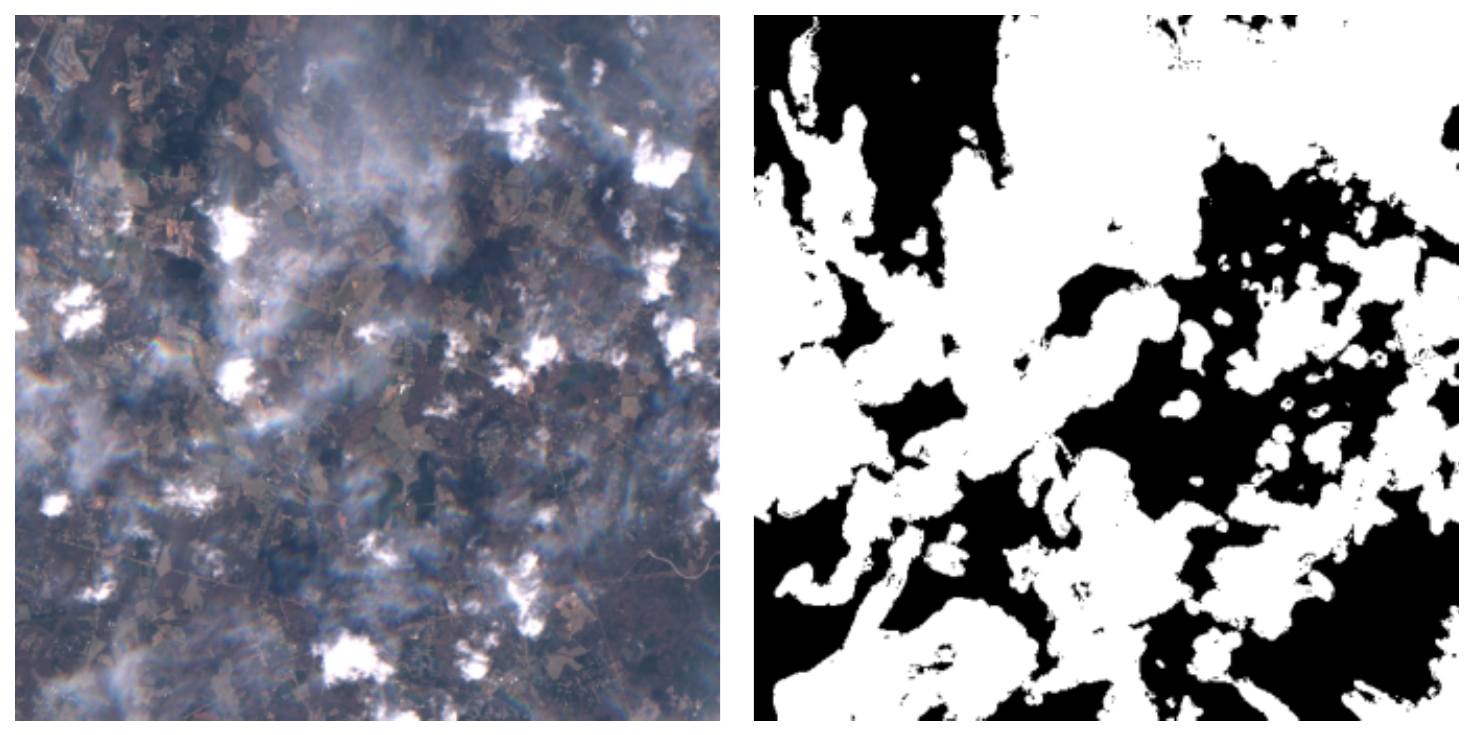}
\caption{Sample from the Sentinel-2 Cloud Mask Catalogue showing an RGB composite (left) and its corresponding ground truth cloud mask (right).}
\label{fig:cloud-dataset-example}
\end{figure}

\begin{figure}[h]
\centering
\includegraphics[width=0.90\linewidth]{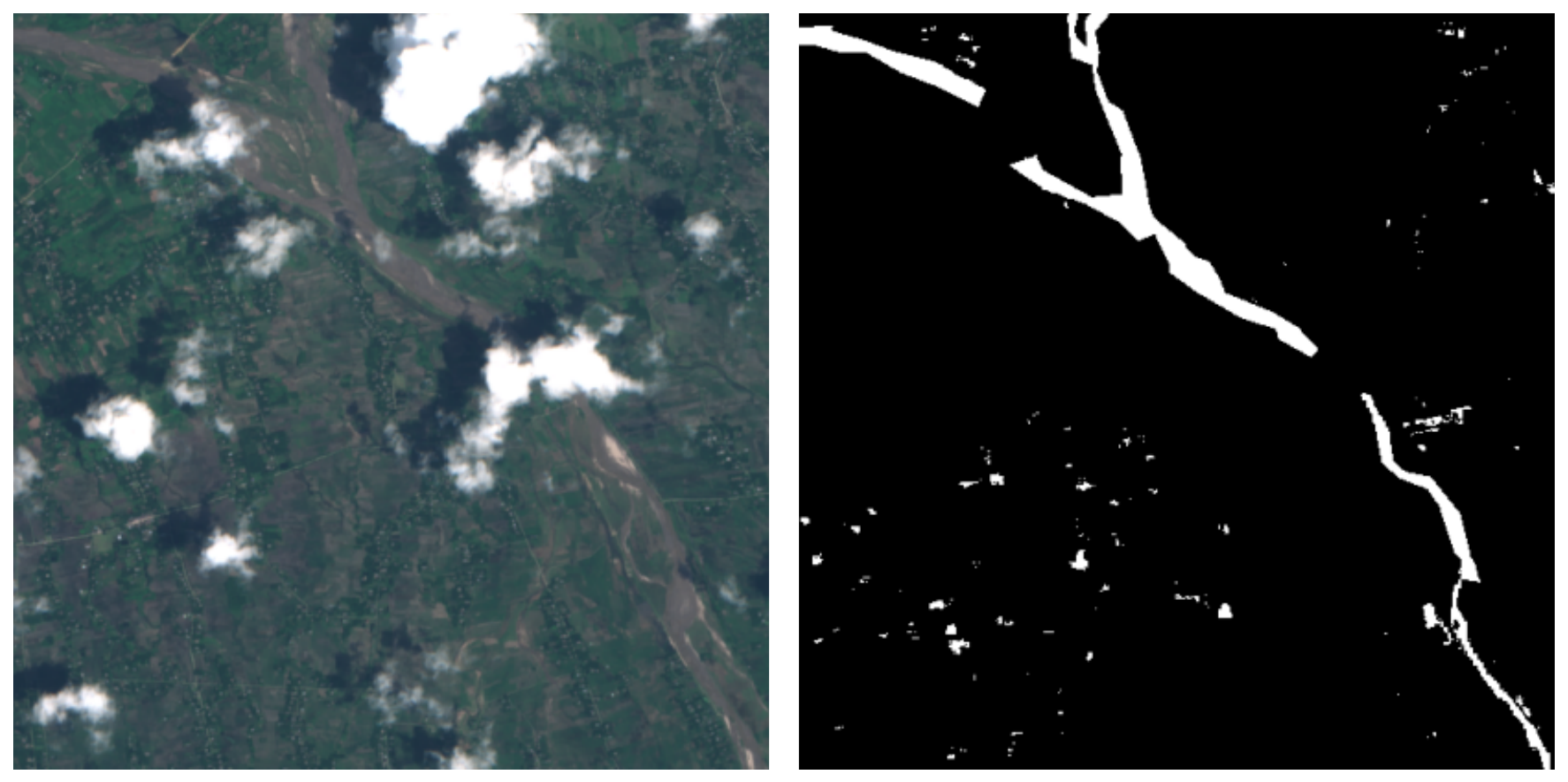}
\caption{Sample from the Sen1Floods11 dataset showing an RGB composite (left) and its corresponding ground truth flood mask (right).}
\label{fig:flood-dataset-example}
\end{figure}

\begin{figure}[h]
\centering
\includegraphics[width=0.90\linewidth]{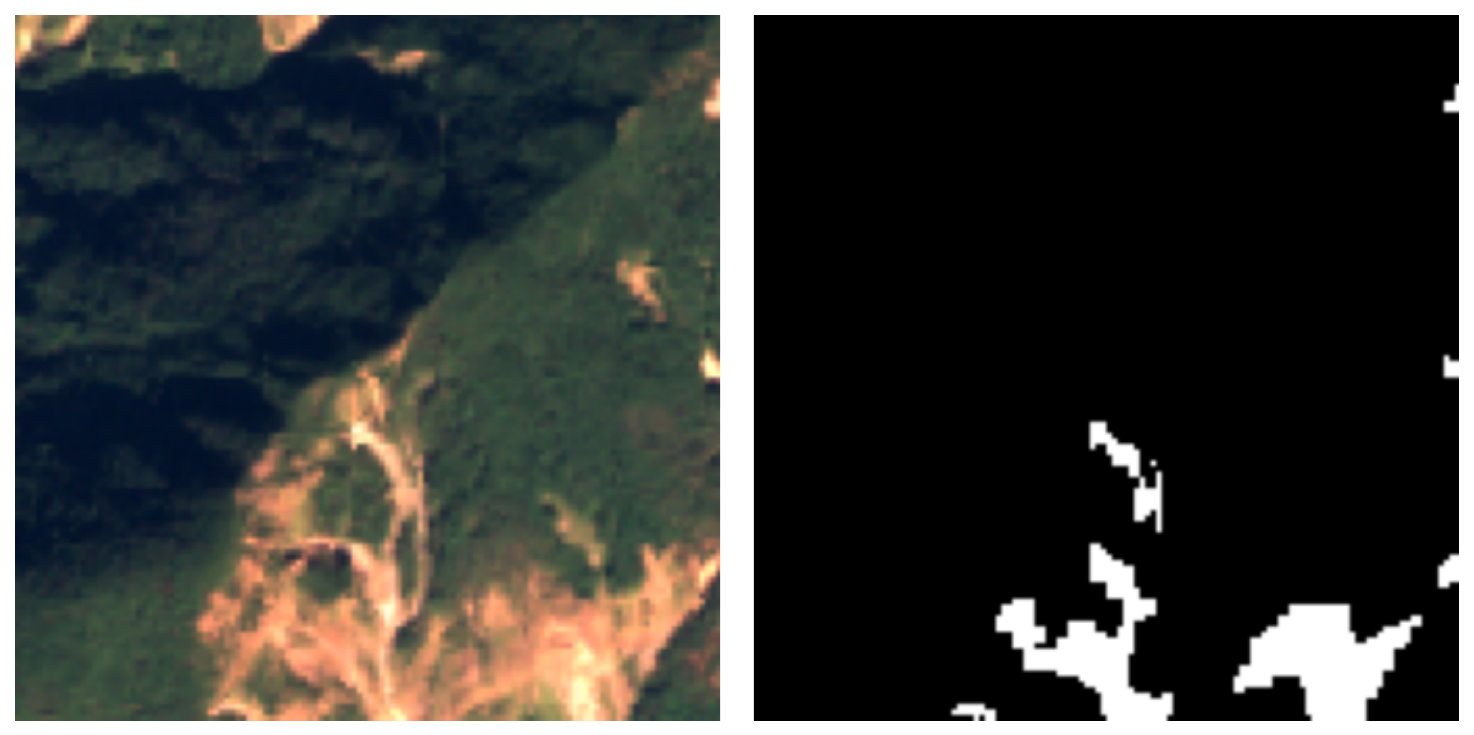}
\caption{Sample from the Landslide4Sense dataset showing an RGB composite (left) and its corresponding ground truth landslide mask (right).}
\label{fig:landslide-dataset-example}
\end{figure}

\begin{figure}[h]
\centering
\includegraphics[width=0.90\linewidth]{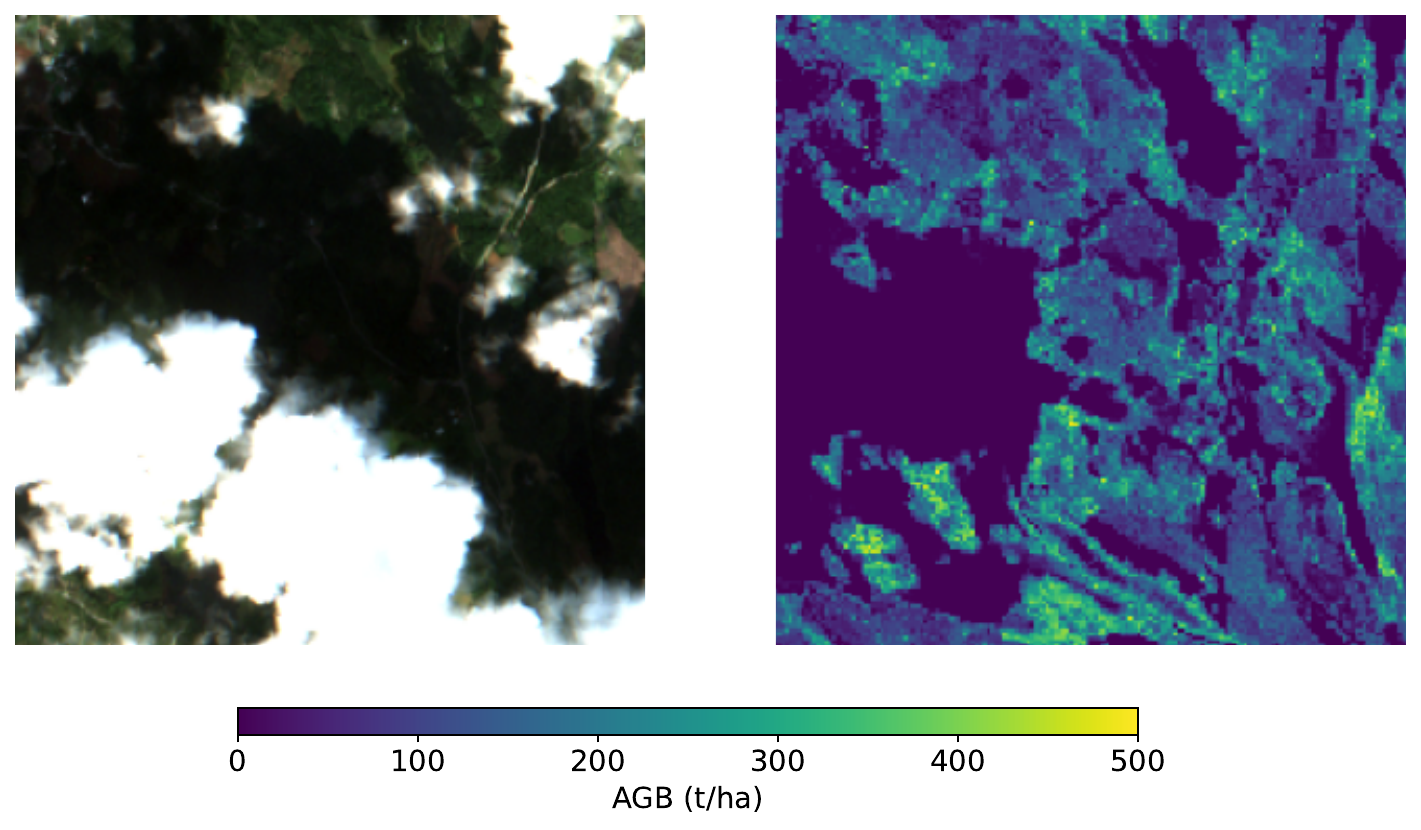}
\caption{Sample from the BioMassters dataset showing an RGB composite (left) and its corresponding AGB ground truth map (right).}
\label{fig:agb-dataset-example}
\end{figure}

\begin{figure}[h]
\centering
\begin{subfigure}[t]{0.85\linewidth}
    \centering
    \includegraphics[width=\linewidth]{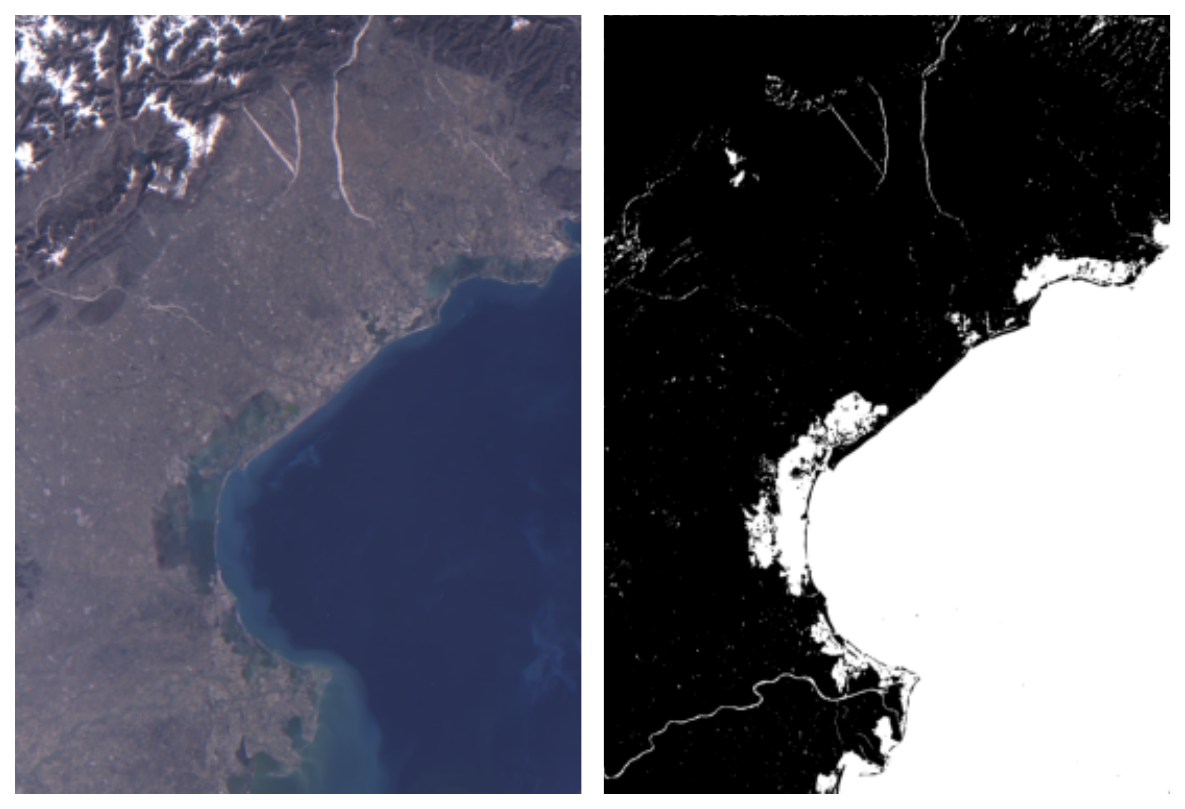}
    \caption{}
    \label{fig:kanyini-water-datacube}
\end{subfigure}


\begin{subfigure}[t]{0.85\linewidth}
    \centering
    \includegraphics[width=\linewidth]{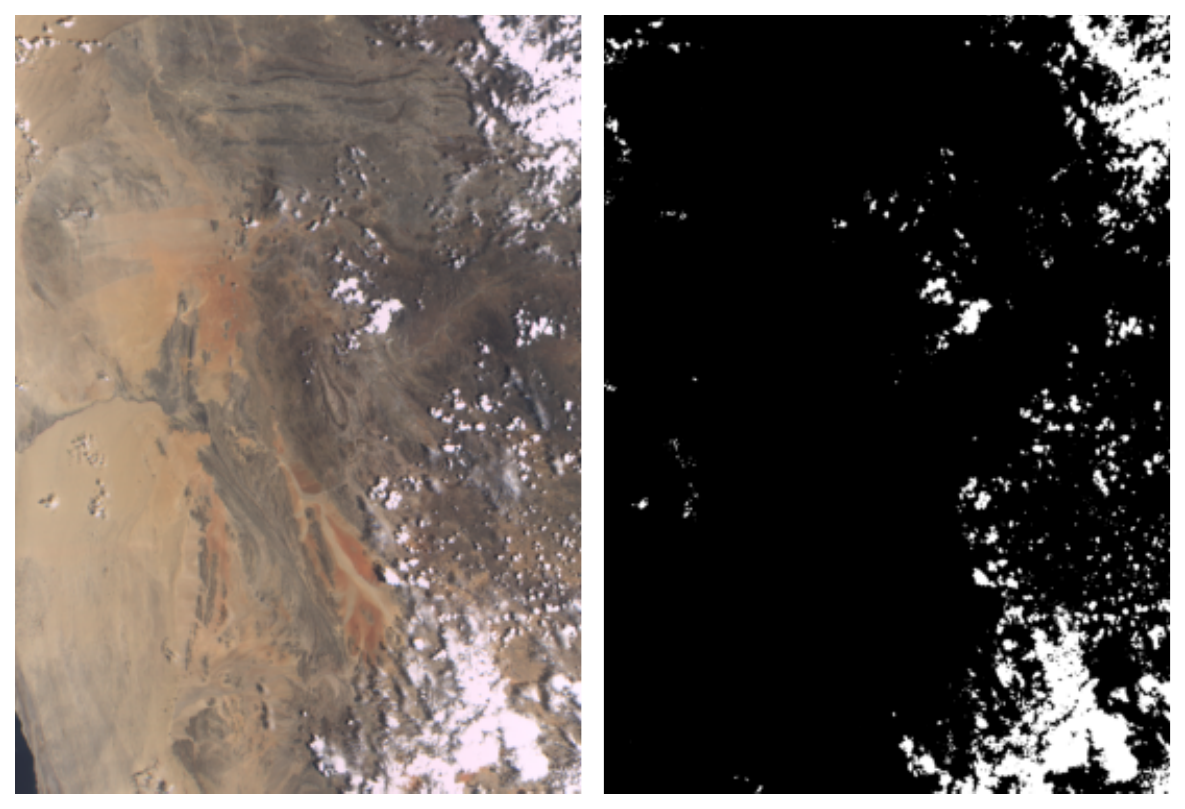}
    \caption{}
    \label{fig:kanyini-cloud-datacube}
\end{subfigure}

\caption{Datacubes captured from Kanyini showing RGB composites (left) and corresponding ground truth masks (right): (a) water and (b) cloud.}
\label{fig:kanyini-water-cloud-datacube}
\end{figure}

\begin{figure}[h]
\centering
\includegraphics[width=\linewidth]{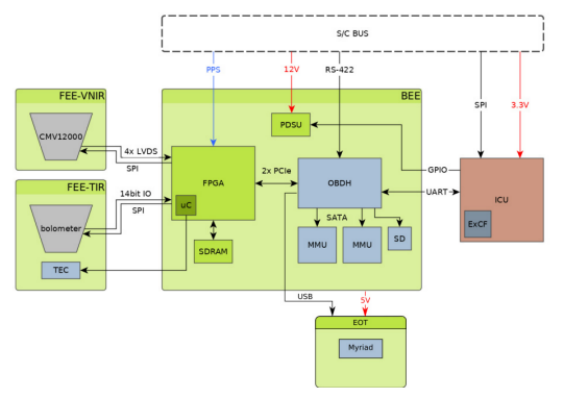}
\caption{Block diagram of the HyperScout-2 payload aboard Kanyini. AI inference on the Myriad-2 is enabled by powering the OBDH, one MMU, and the EOT board.}
\label{fig:hyperscout2-block-diagram}
\end{figure}

\begin{figure}[h]
  \centering
  \begin{subfigure}{0.38\linewidth}
    \centering
    \includegraphics[height=3.3cm]{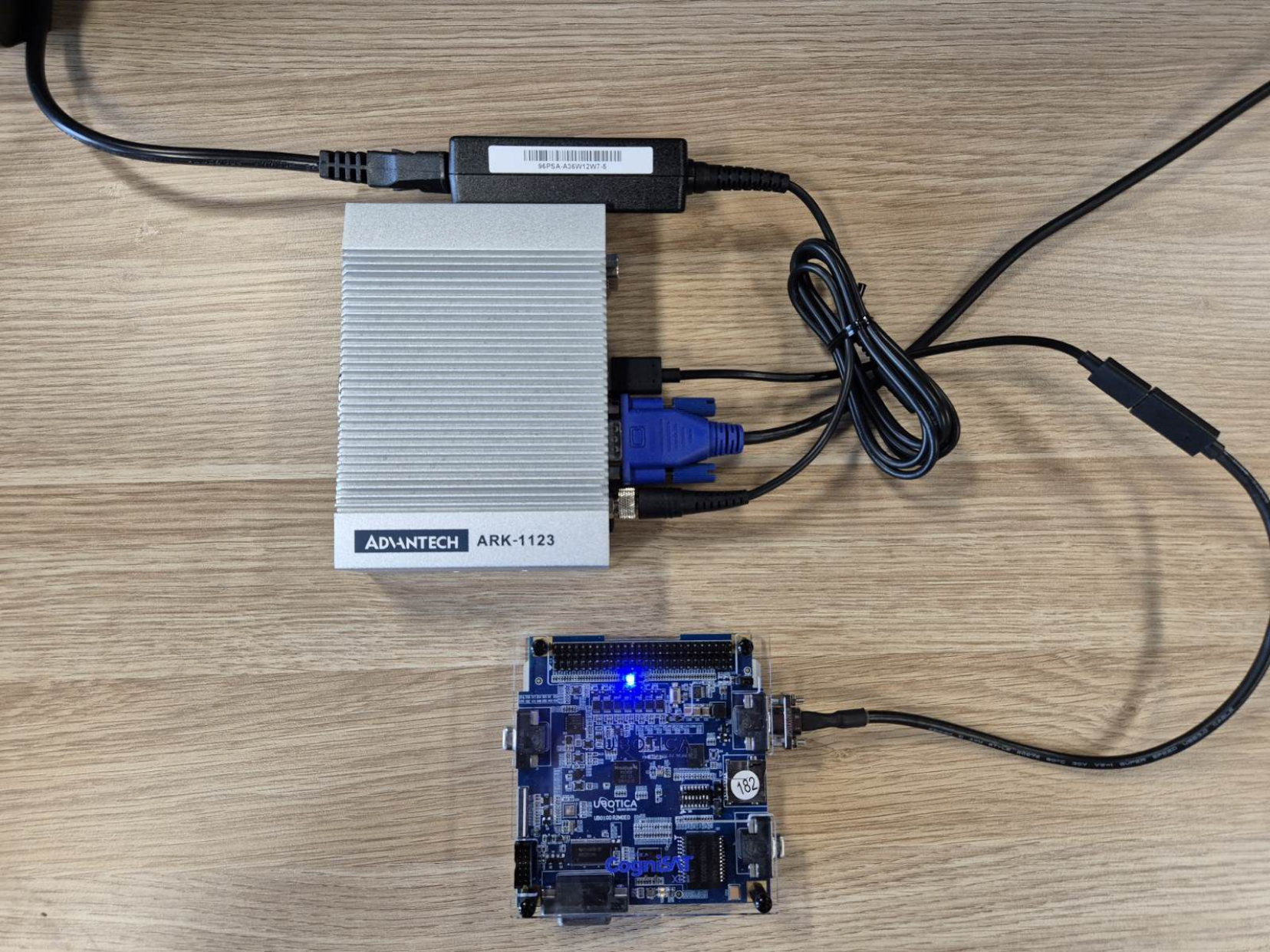}
    \subcaption{\label{fig:hardware-emulator}}
  \end{subfigure}
  \hfill
  \begin{subfigure}{0.58\linewidth}
    \centering
    \includegraphics[height=3.3cm]{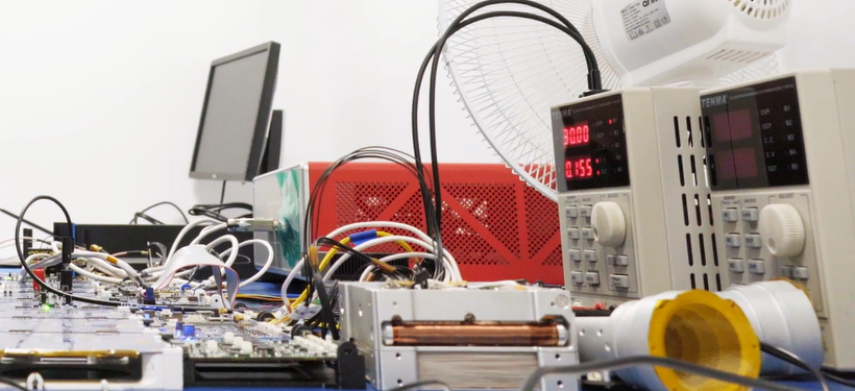}
    \subcaption{\label{fig:engineering-model}}
  \end{subfigure}
  \caption{Ground-based development and testing environments used for the Kanyini mission prior to satellite deployment. The hardware emulator (a) and HyperScout-2 engineering model (red enclosure) (b) provide controlled environments for hardware–software integration, task-level performance, and resource utilisation.}  
  \label{fig:testing-environment}
\end{figure}

\begin{figure}[h]
\centering
\includegraphics[width=0.96\linewidth]{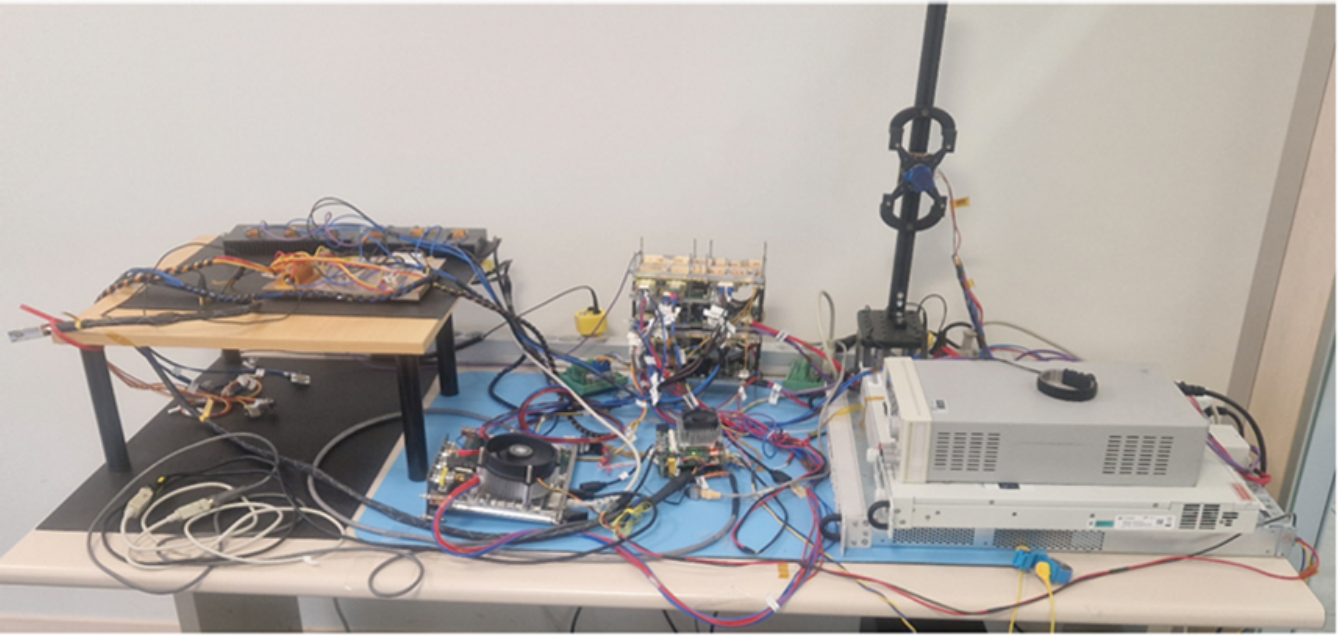}
\vspace{1em}
\caption{Engineering model of the IMAGIN-e compute module used to verify model packaging, containerised deployment, and inference performance before the on-orbit demonstration.}
\label{fig:imagin-e-engineering-model}
\end{figure}


\begin{table}[h]
\centering
\small
\caption{Comparison of Prithvi-V2-EO encoder configurations by patch embedding dimension ($D$), number of weights, and model size in FP32 precision.}
\label{tab:encoder_config}
\begin{tabular}{lccc}
\hline
\textbf{Encoder Configuration} & \textbf{Embed Dim.} & \textbf{\# Weights (M)} & \textbf{Model Size (MB)} \\
\hline
Prithvi-EO-2.0-300M       & 1024      & 303.36       & 1157.03 \\
Prithvi-EO-2.0-512        & 512       & 76.18        & 290.56  \\
Prithvi-EO-2.0-256        & 256       & 19.22        & 73.30   \\
\hline
\end{tabular}
\end{table}

\begin{table}[h]
\centering
\small
\caption{Architectures of task-specific heads for each downstream task, with weight counts and FP32 model sizes. Values vary depending on the selected Prithvi encoder (see Table~\ref{tab:encoder_config}).}
\label{tab:downstream_tasks}
\begin{tabular}{lccc}
\hline
\textbf{Downstream Task} & \textbf{Architecture} & \textbf{\# Weights} & \textbf{Model Size (MB)} \\
\hline
Cloud classification            & 2-layer MLP     & 133k~–~526k   & 0.51~–~2.01 \\
Cloud segmentation              & UNet decoder    & 726k~–~972k   & 2.77~–~3.71 \\
Flood detection                 & UPerNet decoder & 2.87M~–~3.82M & 10.95~–~14.56 \\
Landslide detection             & UNet decoder    & 726k~–~972k   & 2.77~–~3.71 \\
AGB estimation                  & UPerNet decoder & 2.87M~–~3.82M & 10.95~–~14.56 \\
\hline
\end{tabular}
\end{table}